\documentclass[final,twoside]{IEEEtran}

\usepackage[pdftex]{graphicx}
\usepackage{subfig}
\usepackage{array}
\usepackage{color}
\usepackage{amsmath}
\usepackage{amsfonts}
\usepackage{setspace}
\usepackage{amssymb}
\usepackage{textcomp}
\usepackage{algorithm}
\usepackage{algorithmic}
\usepackage[normalem]{ulem}

\usepackage{amsmath}
\usepackage{amsthm}
\usepackage{dsfont}

\newcommand{\captionfonts}{\footnotesize}
\makeatletter  
\long\def\@makecaption#1#2{%
  \vskip\abovecaptionskip
  \sbox\@tempboxa{{\captionfonts #1: #2}}%
  \ifdim \wd\@tempboxa >\hsize
    {\captionfonts #1: #2\par}
  \else
    \hbox to\hsize{\hfil\box\@tempboxa\hfil}%
  \fi
  \vskip\belowcaptionskip}
\makeatother   

\newcommand{\mat}[1]{\mathbf{#1}}
\renewcommand{\vec}[1]{\mathbf{#1}}
\newcommand{\spa}[1]{\mathbb{#1}}

\newcommand{\x}{\vec{x}}
\newcommand{\xs}{\vec{x}^*}
\renewcommand{\c}{\vec{c}}
\renewcommand{\a}{\vec{a}}

\newcommand{\R}{\spa{R}}
\newcommand{\Rv}[1]{\spa{R}^{#1}}
\newcommand{\Rm}[2]{\spa{R}^{#1 \times #2}}

\newcommand{\Id}[1]{\mathds{1}_{#1}}

\newcommand{\cavec}{\begin{pmatrix}\psi_c(\c) \\ \psi_a(\a)\end{pmatrix}}
\newcommand{\cavecT}{\begin{pmatrix}\psi_c(\c) \\ \psi_a(\a)\end{pmatrix}^{\!\!\!T}}

\begin{document}

\title{Where do goals come from? A Generic Approach to Autonomous Goal-System Development}%
\author{Matthias~Rolf and Minoru Asada\thanks{Matthias~Rolf and Minoru Asada are with the Dep. Adaptive Machine Systems at Osaka University, Japan. Mail:~\texttt{\{matthias,asada}@ams.eng.osaka-u.ac.jp\}.}\thanks{This study has been supported by the JSPS Grant-in-Aid
for Specially promoted Research (No. 24000012). 
This research is related to the European Project CODEFROR (PIRSES-2013-612555).
We also would like to thank the Yahoo! Webscope program for providing the “R6B - Yahoo! Front Page Today Module User Click Log Dataset, version 2.0” for our experiments.}}%

\markboth{IEEE Transactions on Autonomous Mental Development (under review)}{Rolf \& Asada: Where do goals come from?  A Generic Approach to Autonomous Goal-System Development}%

\maketitle

\begin{abstract}
Goals express agents' intentions 
and allow them to organize their behavior based on
low-dimensional abstractions of high-dimensional world states.
How can agents develop such goals autonomously? This paper proposes
a detailed conceptual and computational account to this longstanding 
problem.
We argue to consider goals as high-level abstractions of lower-level intention
mechanisms such as rewards and values, and point out that goals
need to be considered alongside with a detection of the own actions' effects.
We propose \emph{Latent Goal Analysis} as a computational learning formulation thereof,
and show constructively that any reward or value function can by 
explained by goals and such self-detection as latent mechanisms.
We first show that learned goals provide a highly effective dimensionality reduction in
a practical reinforcement learning problem.
Then, we investigate a developmental scenario in which 
entirely task-unspecific rewards induced by visual saliency
lead to self and goal representations that constitute goal-directed reaching.
\end{abstract}


\section{Motivation}

Goals are abstractions of high-dimensional world states
that express intelligent agents' intentions underlying their actions. 
Goals are considered to organize the behavior of both humans and robots.
For instance in robot \emph{planning} \cite{bruce2002real} as well as motor \emph{control} 
\cite{HandbookRoboticsControl,Nguyen2011} goals describe the desired outcome of future actions
in terms of what aspect or variable in the world is relevant and what its desired value is.
Also in robot \emph{learning} goals have been proven useful as a scaffolding mechanism 
to perform efficient exploration also in high dimensional sensorimotor spaces \cite{Rolf2010-TAMD,Baranes2013}.
Yet, in all of these scenarios the goals are carefully handcrafted: the variable
to be controlled, as well as 
a mechanism to select a particular goal of the agent at any time
need to be specified by the designer. 
Several formulations of motor learning can
automatically choose internal goals purely for the sake of training a skill (e.g. \cite{Rolf2013-ICDL,Baranes2013}),
but they do neither explain how to choose goals depending on external stimuli, nor 
can they determine the variable that has to be controlled.

Goals are a fundamental concept also in neuroscience and psychology. The entire formulation
of the cerebellum providing \emph{internal} forward and inverse \emph{models} \cite{WolpertMiallKawato1998} only makes sense if goals 
are already given as input for the inverse models.
From a conservative standpoint such models concern motor control in the first place.
Recent theories, however, go much further and suppose that they also contribute to cortex-wide 
higher-level cognitive processes \cite{Ito2008} and are engaged in social behavior \cite{WolpertDoya2003}.
Goals are also seen as a major factor in motivation psychology \cite{locke1994goal}, where 
``goal-setting'' is considered essential for long term behavior organization.
Goals are considered a likewise structuring element in our cognition and perception 
of other agents' behavior such as in imitation learning \cite{bekkering2000imitation} or teleological
action understanding \cite{Gergely2003,Wrede2012-teleological}.

Goals are useful abstractions. But where do they come from?
Neither robotics and machine learning, nor neuroscience, nor psychology provide 
conclusive or even general hints how a biological or artificial agent that starts
with no goals can acquire them.
Already Weng \emph{et al.} \cite{Weng2001} in 2001 pointed out that developmental robots should learn
without a particular task (i.e. explicit goals) given to them at first. But then, how 
could they ever develop such abstractions?
The consequent need to consider the learning \emph{of} goals was explicitly pointed out firstly by Prince \emph{et al.}
\cite{Prince2005} in 2005, but is essentially unsolved ever since. 
It may not be difficult to think of heuristics for an agent to acquire goals within isolated special scenarios,
but what could be general mechanisms for a development of goal systems?
This article seeks for answers to this longstanding question.
We thereby focus on ($i$) the learning of an agent's \emph{own} goals, in contrast to 
observational learning about others' goals such as in imitation learning \cite{schaal1999imitation,muhlig2009automatic}
or  values such as in inverse reinforcement learning \cite{ng2000algorithms}, and ($ii$)
a fully \emph{autonomous} learning without external supervision such as an agent being told what to do.

\subsection{Overview}

Our paper makes four contributions along with the next four sections.
Firstly, we develop a general \emph{conceptualization} of goals in Sec. \ref{sec:goal-def}.
We introduce several novel arguments in order to define goals as precisely as possible,
and argue to consider goal systems as abstractions of reward systems.
These arguments are meant to stimulate a wider discussion and provide the basis for our second contribution:
In Sec. \ref{sec:lga} we introduce a computational learning formulation \emph{Latent Goal Analysis} (LGA).
Based on given rewards we show how ``latent'' goals can be extracted 
from the sensory and action information and constructively prove the universal existence of this transformation.
Thirdly, we show an \emph{application} oriented experiment in Sec. \ref{sec:exp-rec}.
We use LGA as a dimension reduction technique for reinforcement learning. 
In a ``recommender'' scenario online news articles have to be recommended to readers. 
We use goals learned by LGA as compact representations of high-dimensional data that outperform standard methods applied in that field.
Lastly, we show an experimental setup to investigate the human development of goals for the case of reaching in Sec. \ref{sec:exp-sal}.
We show that this task-unspecific information seeking reward based on visual saliency leads to the representation of
to-be-reached objects as goals and a self-detection of the own hand.
We thereby not only learn those abstractions, but already utilize them by applying goal babbling \cite{Rolf2011-ICDL}
to generate actions in a fully bootstrapping and closed action-perception-learning loop.

\section{What is a goal?}\label{sec:goal-def}

How can we operationalize the acquisition of goals? 
In order to achieve a general conceptualization we start from common sense as in dictionary definitions
and relate that meaning to usages in various scientific fields.
Starting from this general definition we distinguish goals from several related terms
such as optimization or affordances.
Most importantly, we point out several vague or entirely unspecified aspects 
in the common sense usage of ``goals'' and make propositions how to substantiate the terminology.
We thereby aim at a conceptualization that is both as general as possible and concrete 
enough for a mathematical operationalization.

\begin{figure}
\begin{center}
\captionsetup[subfigure]{margin=0.5pt,format=hang,parskip=0.5pt,singlelinecheck=true}
\subfloat[][Common sense elements of goals.]{\label{concept1}
\includegraphics[width=0.85\columnwidth,trim = 0cm 0.55cm 0cm 0.2cm, clip=true]{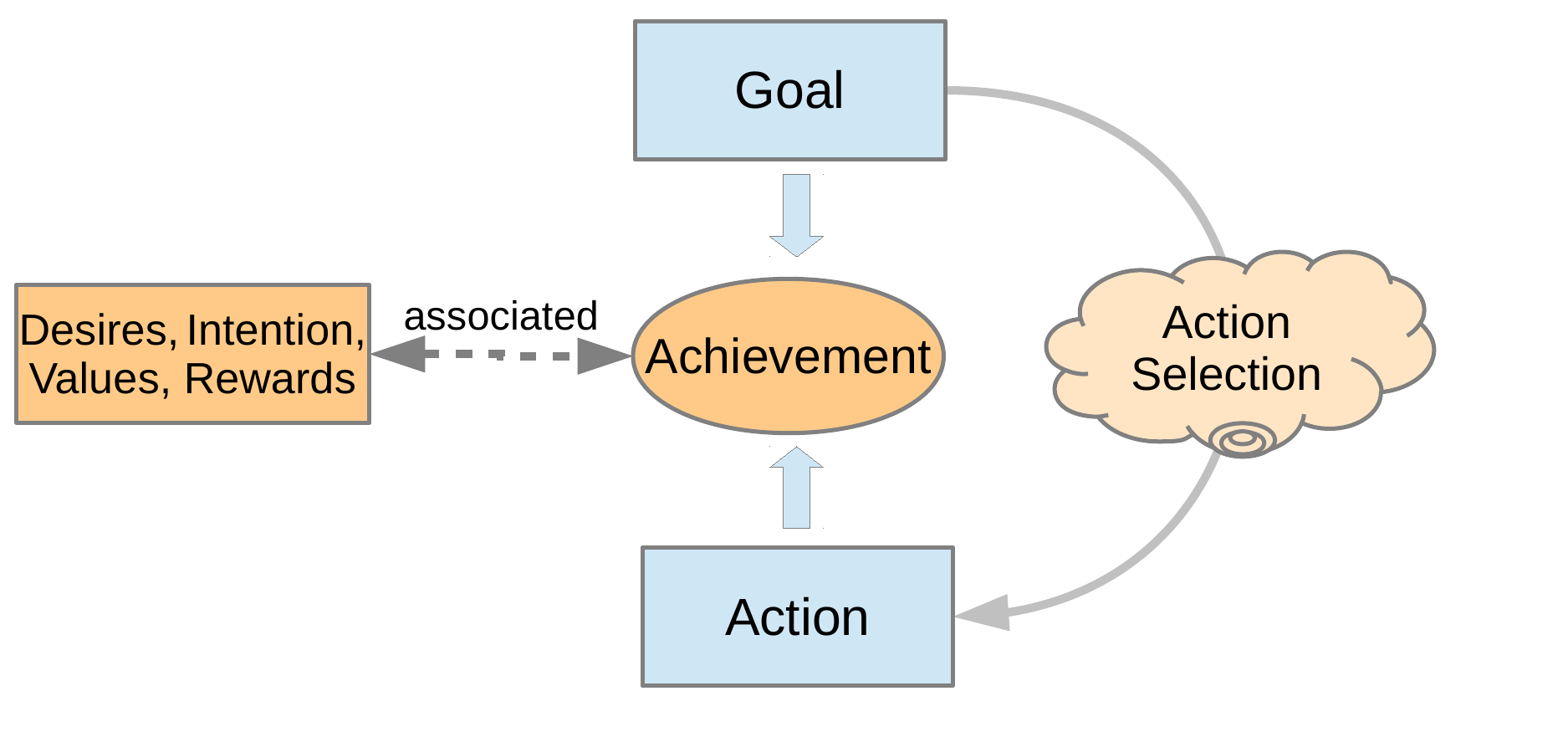}
}\vspace*{-0.2cm}\\
\subfloat[][Proposed conceptual refinement of ``goals''.]{\label{concept2}
\includegraphics[width=0.9\columnwidth]{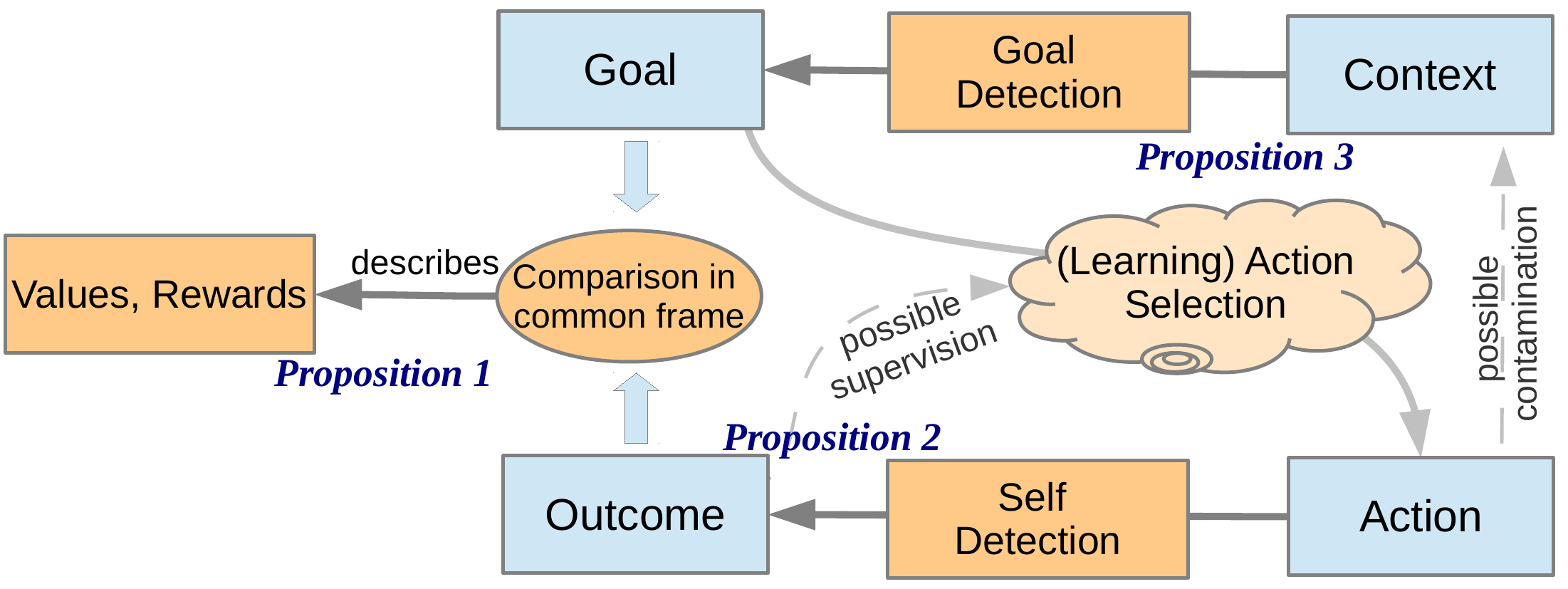}
}
\caption{We argue to consider goals as abstractions of rewards and values.}
\label{concept}
\end{center}
\end{figure} 

\subsection{General Definition and Related Terms}

Dictionaries refer to the term ``goal'' as
``an aim or desired result'' of someone's ``ambition or effort''\footnote{oxforddictionaries.com ``goal''; corresponding definitions in other languages: duden.de (German) ``Ziel'';
nlpwww.nict.go.jp/wn-ja/ (Japanese) Synset 05980875-n; queried 2014/01/15},
Goals are most precisely defined in computational domains that use them.
In motor coordination and control \cite{HandbookRoboticsControl,Nguyen2011} goals are typically low-dimensional abstractions 
of the task such as a desired angular velocity of an electric motor
or a desired position of a robot's end effector. Thereby goals are formulated as \emph{values} in some 
\emph{low-dimensional space} (e.g. 1d velocities, 3d positions) in which they abstract from many task-irrelevant variables 
(such as room temperature) of the much higher-dimensional physical processes.
Also goals in planning \cite{bruce2002real} describe world variables 
that should have some desired value, while other variables are irrelevant. 
In both control and planning the goal has to be \emph{achieved} by means of \emph{action}, i.e.
the the agent has to find actions that result in the observation of the desired variable values.
Similar aspects can also be found in goal-setting psychology \cite{locke1994goal}:
For instance in management psychology it has been proposed that goals should be specific (have a particular value), measurable, 
and realistic (i.e. achievable by means of own action) \cite{Doran1981-SMART}.
The above points clearly distinguish goals from two other kinds of desires or intentions:
($i$) \emph{Optimization} or general improvement (such as increasing reward) are not goals in a narrow sense.
``Improvement'' for itself is not specific in the sense that a particular to be achieved value is specified.
Hence, there is no definite achievement possible or an end defined.
($ii$)~\emph{Wishes} of desired world states (e.g. having a sunny say)
are no goals because they are not achievable by means of own action in the first place.

With these aspects we can attempt a first definition that we will refine in the remainder of this section:
\begin{quote}
\emph{Definition}: A \emph{goal} is an equivalence set of world states that, in a certain situation, 
an agent desires to achieve as a result of its own action.
\end{quote}
They refer to an equivalence set of states in the sense that there can be irrelevant variables 
that to not matter for the goal. Hence, any of their values are equivalently acceptable.
The main point is that goals reflect a particular desire. 
Their achievement has some value to the agent.
This stands in contrast e.g. to
\emph{affordances} \cite{Gibson1977}. Sahin \emph{et al.} formulated affordances as the relation 
between the action of an agent, an object under manipulation, and the effect on that object~\cite{Sahin2007}.
Objects with similar action-effect relations can then be summarized in equivalence classes 
such as ``standonable''. Related to this formulation, \emph{contingencies} \cite{regan2001contingency,nagai2012contingency} 
and action-effect bindings \cite{Hommel2009,Verschoor2009} 
describe general relations between actions and their specific effects.
Goals and affordances are both interactivist 
concepts that can not be defined by only the agent or only the environment, but only via their interplay.
The crucial difference between them is that affordances describe any \emph{possible}
thing that could be done. Affordances are not associated to any value or \emph{desire}, while 
this desire to do something is the constituting aspect of goals in our view.

\subsection{A Refined Concept of Goals}

Intelligent organisms can not arbitrarily regard things as goals while others are not regarded as goals.
The biggest open question about goals in our view is \emph{what kind of information could lead to this differentiation
between goals and non-goals}.
This source of information, or learning signal, is consequently the main point of our overall argumentation.
In terms of machine learning we know three basic kinds of learning signals:
\emph{supervised} input of ground-truth values, \emph{unsupervised} learning of input statistics,
and \emph{reward} or cost signals in reinforcement learning or optimization.
Supervised learning as source of information seems entirely unsuited for autonomous 
development of goals, since a teacher for such information would have to be external.
While social learning of goals in such terms certainly exists, it does not provide 
answers for an ontogenetic core mechanism of goal systems development.
Unsupervised learning seems likewise unsuited since signal statistics 
can not tell about a desire or value. 
Rather, reward signals seem to be the suitable learning signal, as they express
the most primitive form of a value. 
Considering goals as high-level abstractions of intention
therefore suggests to consider them \emph{abstractions of} world states that are associated to reward:
\begin{quote}
\emph{Proposition 1}: Goals are \emph{abstractions} that do not themselves \emph{determine} a desire,
but rather \emph{describe} it based on lower-level systems of desire, such as reward or value systems.
\end{quote}
Corresponding rewards might reflect some task very directly when e.g. determined in a social context, or directly as food.
However, they might also be purely \emph{internal} or \emph{intrinsic} as it is often considered 
in the contexts of intrinsic \emph{motivation} \cite{ryan2000intrinsic,Schmidhuber2010,Oudeyer2007,Baldassarre2011} or
information seeking \cite{Gottlieb2012}. Our exemplary experiment in Sec. \ref{sec:exp-sal} will 
take the latter perspective.
The abstraction process thereby could not only concern immediate rewards, but also expected, future rewards that are expressed in \emph{value systems} (e.g. supposed to exist in the midbrain \cite{Schultz1997}).
In Fig. \ref{concept2} we illustrate this proposition by saying the achievement of a goal (the abstraction) ``describes''
an (expected) reward. 

The second question we must clarify to conceptualize goals is \emph{what ``achievement'' means}.
In natural language it is often expressed that an ``action achieves a goal''. However,
this skips an important aspect that must be considered for a profound conceptualization:
Goals do not come alone, but always paired with an \emph{evaluation} of the own action's effect.
It is the \emph{effect} of the action, not the action itself to which the goal is compared.
In robot reaching this evaluation, or rather its learning, is often referred to as 
\emph{self-detection}
\cite{edsinger2006can,Stoytchev2011}: the robot's hand needs to be detected for instance in a camera 
image.
Just in the case of reaching, this notion is related from ``body schema'' \cite{hoffmann2010body} that describe a localization of the body in general.
Here we use the term self-detection also in a non-reaching and non-spatial sense that describes any effect related to a goal.
For instance the goal of having a sandwich to eat (e.g. followed by the verbal action ``make me a sandwich'') 
would be accompanied by the detection of actually having a sandwich or not.
The evaluation of the action equally exists in planning (e.g. navigation) domains,
where the effect of past action is compared to the desired outcome,
and goal-setting psychology, where a major emphasis is made on measurability of goals.
In all of these domains the effect of the own action (self-detection) is then compared to the goal within a common
reference frame in order to assess the achievement.
The need for this comparison forbids considering the development of goals and self-detection separately from each other:
\begin{quote}
\emph{Proposition 2}: Goals cannot be learned or considered independently from \emph{self-detection}, but both have
to describe a consistent \emph{reference frame} in which the goal can be compared to the outcome of an action.
\end{quote}
This aspect will largely guide our computational formalization. Our experiments will also 
illustrate that this aspect poses an important developmental hallmark: the entrance of \emph{self-supervised} 
action and motor learning. Once self-detection and goals are available, a supervised learning signal
becomes available to other learning processes. When self-detection (e.g. a robot's forward function)
and goals were already available they have been used in numerous approaches for motor learning 
already \cite{Nguyen2011}. With respect to the autonomous development of goals already
Prince \cite{Prince2005} noted that goals are related to self-supervision, but missed the point that 
it is not the goals themselves, but rather the self-detection (in relation to the goals) that enables 
the supervision.

Finally, we need to consider how goals become ``active'', i.e. how an agent determines
which goal to follow at a present moment. It is often considered that agents can have 
multiple goals, e.g. on different timescales or also parallel or secondary goals in the long run.
This leaves a lot of play for an operationalization, which we propose to organize
with the following restriction using the notion of cognitive 
or processing ``(sub-)systems'' internal to an agent:
\begin{quote}
\emph{Proposition 3}: One system can have only \emph{one} (active) goal \emph{at a time}, which expresses the \emph{present} desire
based on an internal or sensory \emph{context}.
\end{quote}
Hence, different systems of cognitive processing (e.g. organizing different timescales of behavior) 
or motor planning and control can have one present goal each. 
Further we note that a goal gets triggered by a \emph{context}, such as sensory information, an internal state
or information from other processing systems.
This seems trivial but makes an important point: there needs to be a \emph{goal-detection} 
to determine the now to be followed goal from the context, and that parallels the self-detection.
In the reaching example this might be a mechanism to determine the position of 
the relevant object from camera images, or also hard-wired position selectors in
many robot setups. In terms of motivational psychology, for instance a certain context 
of a conversation might trigger a subsequent communicational goal such as to convey an information
(``by the way...'').

In summary, we refer to a \emph{goal system} as the joint apparatus of goal-detection from a context and self-detection
that are compared within a common reference frame, such that the achievement of the goal 
by means of own action is reflected by a reward or value (see Fig. \ref{concept2}).

\section{Latent Goal Analysis}\label{sec:lga}

\begin{figure*}
\begin{center}
\captionsetup[subfigure]{margin=0.5pt,format=hang,parskip=0.5pt,singlelinecheck=true}
\subfloat[][Motor control with internal models.]{\label{spaces-control}
\includegraphics[height=3.0cm,trim = 0cm 0.45cm 0cm 0.1cm, clip=true]{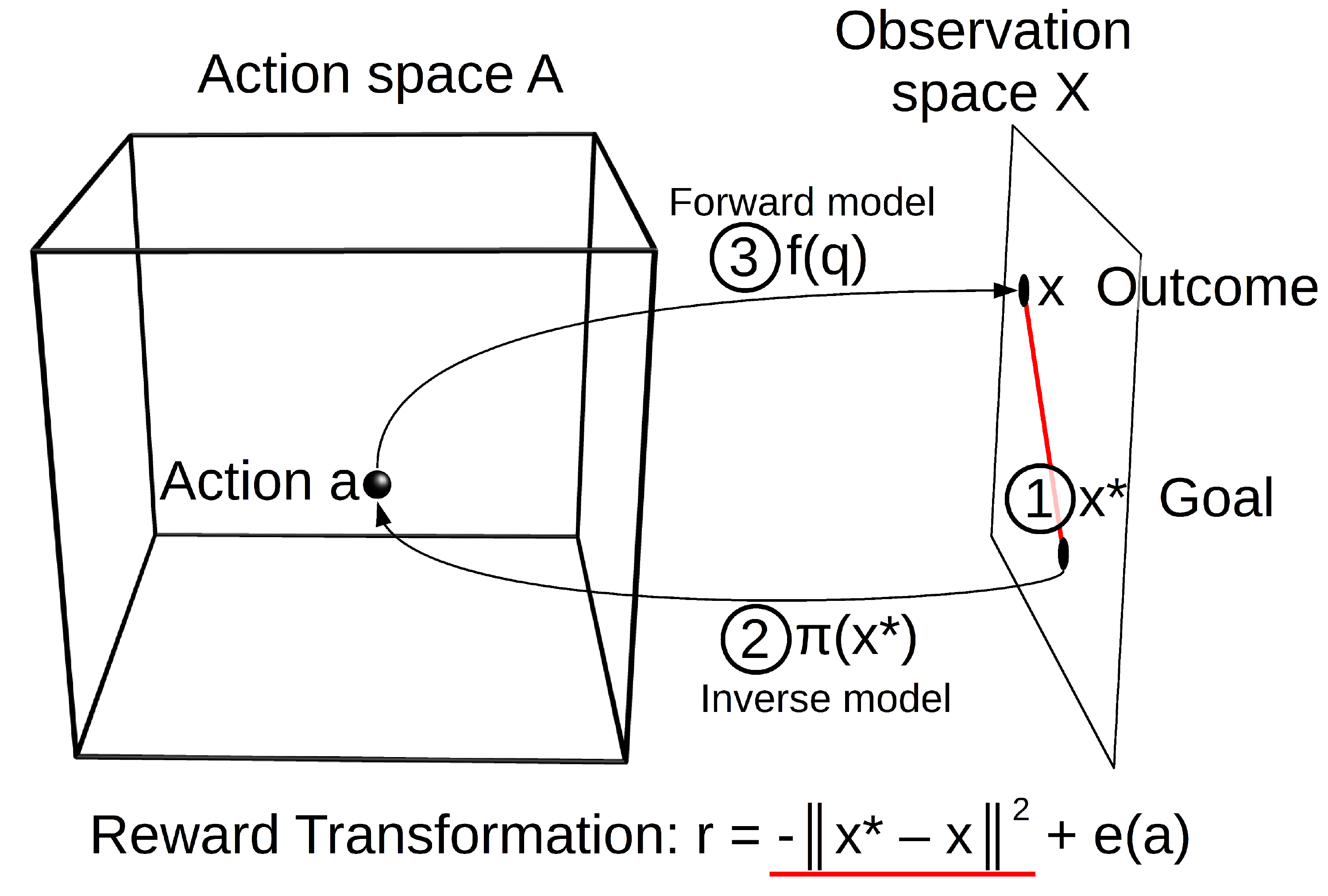} 
}
\subfloat[][One-step reinforcement learning.]{\label{spaces-reward}
\includegraphics[height=3.0cm]{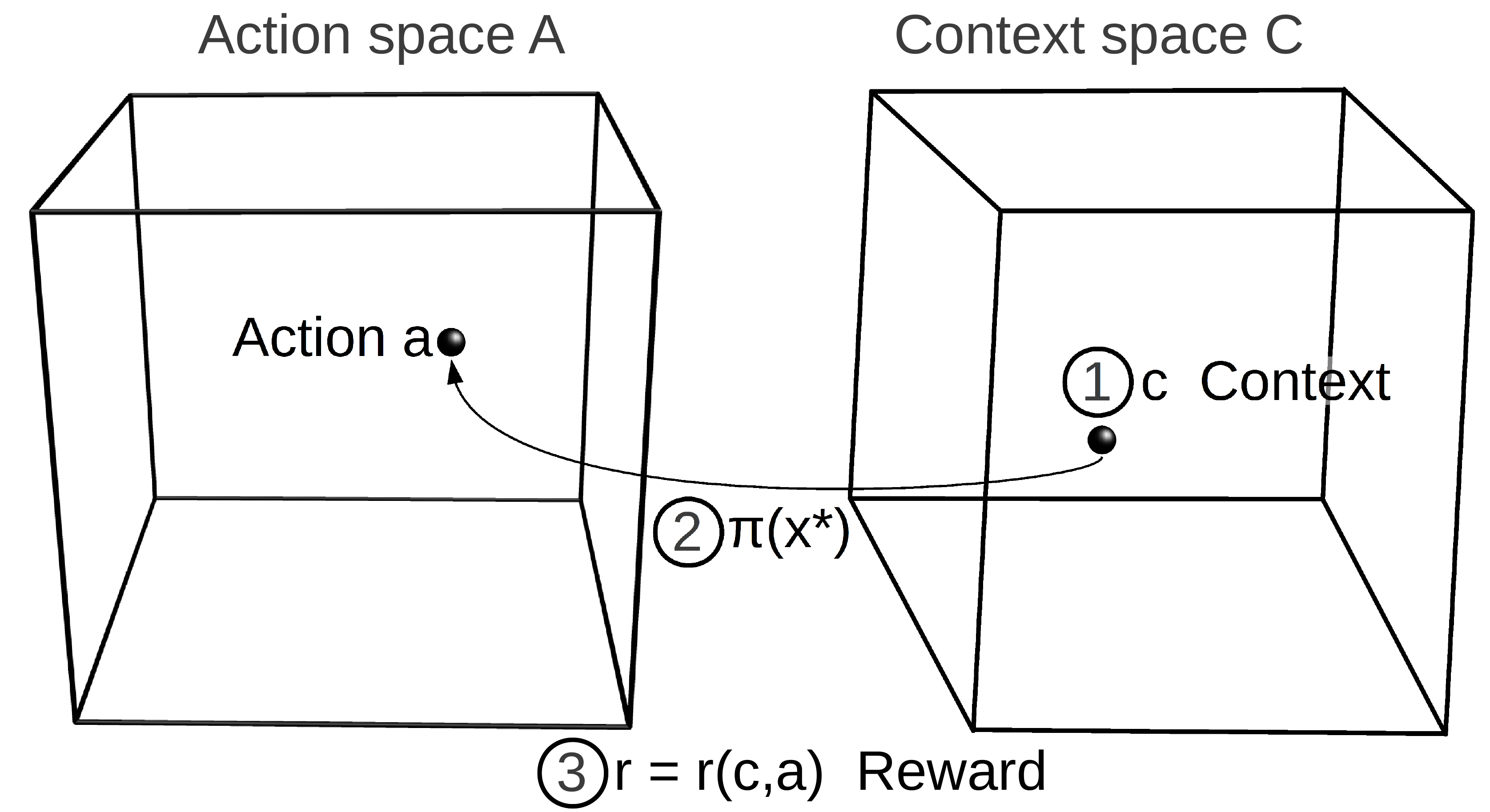}
}
\subfloat[][Latent goal analysis (LGA).]{\label{spaces-lga-dec}
\includegraphics[height=2.8cm]{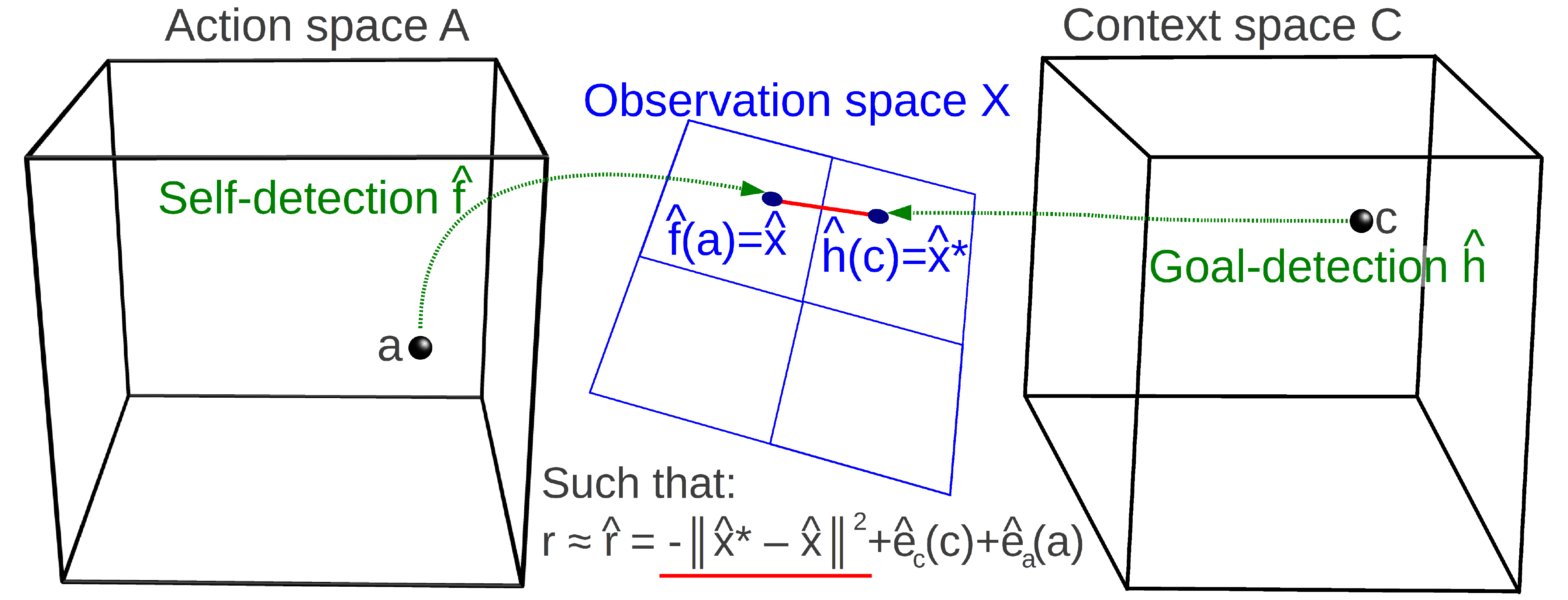}
} 
\caption{Latent goal analysis (LGA) identifies how to project actions and contexts into a common observation space.
	  The observed rewards $r$ are thereby explained by distance between 
	  action-outcome $\x$ (self-detection) and goal $\xs$ (goal-detection),
	  such that a reward problem is turned into a control problem.}
\label{spaces-lga}
\end{center}
\end{figure*} 

In the last section we have established a conceptual framework of goals and argued to learn 
them together with self-detection as abstractions of reward or value signals.
In order to develop a mathematical learning framework, we now have to set those terms in a \emph{formal} relation.
The best way to do so is to consider a domain which \emph{already} has a 
formal relation of at least most of the conceptual elements, such as \emph{control} or \emph{coordination} problems
\cite{HandbookRoboticsControl,Nguyen2011,Rolf2010-TAMD}. In this section we develop our computational
learning framework for goals by introducing this set of formal relations, adding terms from the concept 
that are still missing (Sec. \ref{sec:rew-trans}), and then establishing a learning rule from rewards to mathematical terms
corresponding to each of the conceptual terms (Sec. \ref{sec:dec}). In order to establish a generic
framework we will not make any assumptions about the source of rewards, but will show examples
for extrinsic (Sec. \ref{sec:exp-rec}) and intrinsic (Sec. \ref{sec:exp-sal}) rewards in the experiments.

It can be argued that starting from such a particular domain as motor control
considerably narrows the scope of the overall concept. However, we will show in this section 
the \emph{universality} of this mathematical approach, i.e. that it can learn goals from \emph{any
reward or value function}. Specifically, we show an equivalence relation between motor control 
and reinforcement learning, such that any reinforcement learning problem can be transformed into
a motor control problem with its abstractions for goals and self-detection. Hence, our learning 
framework is as general as reinforcement learning, which is often considered the most general 
learning formulation at all.

\subsection{Reward Transformation}\label{sec:rew-trans}

We start by establishing a formal relation of the conceptual terms based on motor control
or coordination problems used in contexts of robot control \cite{HandbookRoboticsControl}
and learning based on internal models \cite{WolpertMiallKawato1998,Nguyen2011,Rolf2011-ICDL,Baranes2013}.
Motor control problems as shown in Fig.\ \ref{spaces-lga}\subref{spaces-control} follow a simple protocol:
$(1)$: The world provides a \emph{goal} $\xs$ to the agent that is situated in some \emph{observation space} $\spa{X}\!\subseteq\!\Rv{n}$. 
$(2)$: The agent chooses an \emph{action} $\a$ from some \emph{action space} $\spa{A}\!\subseteq\!\Rv{m}$.
$(3)$: The world provides a causal \emph{outcome} $f(\a)\!=\!\x$ of the agent's action, again situated in $\spa{X}$. 
The agent's task is to choose an action such that the outcome $\x$ matches the goal $\xs$: $\x\!=\!f(\a)\!=\!\xs$.
Many coordination problems provide \emph{redundancy}: the action space is substantially higher dimensional than
the observation space ($n\ll{}m$), such that multiple actions $\a_i\!\neq{}\!\a_j$ map to same outcome $f(\a_i)\!=\!f(\a_j)$. 
In such scenarios 
additional cost functions $-e_a(\a)$ are often considered \cite{HandbookRoboticsControl} to select an optimal action among all those that fulfill $f(\a)\!=\!\xs$.
In the case of hand-eye coordination $\a$ could correspond to joint angles or torques. $\x\!=\!f(\a)$ is the position  of the end-effector that results from applying $\a$. 
The ground truth function $f\!:\spa{A}\!\rightarrow{}\!\spa{X}$ is called \emph{forward function}. 
When this function is learned from supervised examples it is called \emph{forward model}.
The task to identify it (without supervision) is called \emph{self-detection} \cite{Stoytchev2011}.
The agent's selection of an action is addressed with an \emph{inverse model} (see Fig.\ \ref{spaces-lga}\subref{spaces-control}) that 
is an inverse function of the forward model \cite{WolpertMiallKawato1998,Rolf2013-NeuCom}.
This problem is often considered by means of an overall cost-function of the 
distance of goal and outcome, and $e_a(\a)$, which is easily transformed into reward
semantics by inverting the sign:
\begin{equation}
 r(\xs,\a) = -|| \xs - f(\a) || ^2 + e_a(\a)
\end{equation}
What is still missing in this formulation is the goal-detection. This mechanism
is not usually part of academic papers about motor control, yet always present:
The goals of control processes are chosen by vision processes identifying relevant objects to 
be manipulated, planning- or other processes that are usually hard-wired.
Hence, they are in some way determined by a larger internal or external context.
We can denote this selection on an abstract level with a function $h(\c)$, which we refer to as \emph{goal-detection}.
Further, we introduce for the sake of symmetry a virtual cost term $e_c(\c)$ that only
depends on the context, and which we will later need for our theoretical considerations. 
This term does not influence the optimal action selection
for a given context, but introduces the aspect that the optimally possible reward 
depends on the context. Altogether this gives the reward transformation
\begin{equation}
 r(\c,\a) = -|| h(\c) - f(\a) || ^2 + e_c(\c) + e_a(\a) \ . \label{eqn:rew-transform}
\end{equation}
The overall protocol now corresponds to a one-step \emph{reinforcement problem} \cite{ML-ASSO-RL,Langford2008} as shown in Fig.\ \ref{spaces-lga}\subref{spaces-reward}:
$(1)$: The world provides a \emph{context} $\c$ in some context space $\spa{C}\subseteq\Rv{p}$.
$(2)$: The agent chooses an \emph{action} $\a$ from the \emph{action space} $\spa{A}\subseteq\Rv{m}$.
$(3)$: The world provides a \emph{reward} $r\in{}\R$ based on latent goals and action outcomes 
	as in equation \ref{eqn:rew-transform}.

\subsection{Latent Goal Transformation}\label{sec:dec}

We have now established a formal relation between all terms in our concept.
Yet, so far this formulation only allows to transform existing goals into rewards.
The central idea for computational learning of goals is now to \emph{invert} this process.
Given \emph{any possible} reward or value function, is it possible to transform in \emph{back} 
into a motor control problem? 

In fact, this should not be possible according to common beliefs about reinforcement learning:
Motor control comes with highly 
informative and usually low-dimensional abstractions that can be used for supervised learning \cite{WolpertMiallKawato1998,Nguyen2011}.
Goal and actual values in motor control define a relation similar (see \cite{Rolf2013-NeuCom}) to actual and 
target outputs in classical supervised learning setups by providing ``directional information'' 
in contrast to a mere ``magnitude of an error'' in reinforcement learning \cite{Barto1994-RLMC}. 
Given this rich structure in motor control, reinforcement learning
seems the by far more general setup \cite{kaelbling1996reinforcement,Barto2004-RL-Supervised}.

Yet, we now show that in fact any reinforcement problem can be turned into a motor control 
problem, and thereby goals and self-detection can be retrieved as abstractions from a reward signal.
Therefore we need to 
find functions $\hat{f}$, $\hat{h}$, $\hat{e}_c$ and $\hat{e}_a$ 
to resemble any possible reward function~$r(\c,\a)$:
\begin{equation}
 r(\c,\a) = \hat{r}(\c,\a) = -|| \hat{h}(\c) - \hat{f}(\a) ||^2 + \hat{e}_c(\c) + \hat{e}_a(\a) \label{eqn:rew-transform-est} \ ,
\end{equation}
or value function $Q(\c,\a)$ expressing expected future rewards:
\begin{equation*}
 Q(\c,\a) = \hat{Q}(\c,\a) = -|| \hat{h}(\c) - \hat{f}(\a) ||^2 + \hat{e}_c(\c) + \hat{e}_a(\a)
\end{equation*}
This work does \emph{not} tackle the temporal credit assignment problem to estimate 
$Q$ itself. However, if a value system \cite{Schultz1997,daw2006computational} to estimate future rewards $Q$ is already
available, decomposing either $r(\c,\a)$ or $Q(\c,\a)$ is computationally equivalent since both 
are scalar functions of $\c$ and $\a$.

The major challenge is to identify the forward model by self-detection $\hat{f}(\a)\!=\!\hat{\x}$, and the goal-detection $\hat{h}(\c)\!=\!\hat{\x}^*$. Thereby goals and outcomes are considered \emph{latent variables} of the reward function.
These abstractions constitute the control problem
by describing the interaction of goals and outcomes in a low-dimensional 
observation space (see Fig.\ \ref{spaces-lga}\subref{spaces-lga-dec}). 
Cost terms depending on context \emph{or} action only are considered as remainders,
and in fact are easy to find given $\hat{f}$ and $\hat{h}$.

\paragraph{Ansatz}

Finding such functions can be formulated as finding appropriate coefficients
of parametrized functions. First, we consider features $\psi_c(\c)\!:\spa{C}\!\rightarrow{}\!\Rv{p'}$ 
and $\psi_a(\a)\!:\spa{A}\!\rightarrow{}\!\Rv{m'}$ to describe the contexts and actions.
Assuming an $n$-dimensional observation space $\spa{X}$ we can denote our function candidates with coefficients $\mat{M}$, $\mat{H}$, $\mat{R}_a$, and $\mat{R}_c$ as:
\begin{eqnarray*}
 \hat{\x}^* = \hat{h}(\c) &=& \mat{H}\cdot{}\Psi_c(\c) \ , \ \ \mat{H}\in\Rm{n}{p'} \\
 \hat{\x} = \hat{f}(\a) &=& \mat{M}\cdot{}\Psi_a(\a) \ , \ \ \mat{M}\in\Rm{n}{m'} \\
 \hat{e}_c(\c) &=& \Psi_c(\c)^T\cdot{}\mat{R}_c\cdot{}\Psi_c(\c) \ , \ \ \mat{R}_c\in\Rm{p'}{p'} \\
 \hat{e}_a(\a) &=& \Psi_a(\a)^T\cdot{}\mat{R}_a\cdot{}\Psi_a(\a) \ , \ \ \mat{R}_a\in\Rm{m'}{m'} \ .
\end{eqnarray*}
When we insert these definitions into Eqn. \ref{eqn:rew-transform-est} we can write the reward transformation of a control problem in matrix notation
\begin{equation}
 \hat{r}(\c,\a) \!\!=\!\!  \cavecT \! \! \! \begin{pmatrix}
             \mat{R}_c \text{-} \mat{H}^T\mat{H} \!\!&\!\! \mat{H}^T\mat{M} \\
             \mat{M}^T\mat{H} \!\!&\!\! \mat{R}_a \text{-} \mat{M}^T\mat{H}
             \end{pmatrix} \! \! \cavec \label{eqn:rew-lga}
\end{equation}
as a quadratic form of context- and action-features.

\paragraph{Observation Space Reconstruction}
We can now write the \emph{actual} reward function $r(\c,\a)$ in a similar form by
\begin{eqnarray}
 r(\c,\a) &=&  \cavecT \! \cdot{} \mat{K} \cdot{} \cavec \label{caKca} \nonumber\\
 &=&  \cavecT \!  \! \begin{pmatrix}
             \mat{K}_{c,c} & \mat{K}_{c,a} \\
             \mat{K}_{c,a}^T & \mat{K}_{a,a}
             \end{pmatrix} \! \cavec . \label{eqn:quadratic}
\end{eqnarray}
This form with a symmetric coefficient matrix $\mat{K}$ is a \emph{universal approximator}: 
it can arbitrarily well approximate at least all continuous functions if appropriate
features $\psi$ are chosen. For instance, if the features $\psi_a$ and $\psi_c$ are separate
polynomial features of $\a$ and $\c$ up to polynomial degree $d$, then just the 
subterm $\psi_c(\c)^T\cdot{}\mat{K}_{c,a}\cdot{}\psi_a(\a)$ will contain all joint polynomial terms of $\a$ and $\c$ up to degree $d$. Hence, equation \ref{eqn:quadratic} can at least describe 
all functions that can be described by polynomials, 
which are at least all continuous functions.

We can now find coefficients $\mat{M}$, $\mat{H}$, $\mat{R}_a$, and $\mat{R}_c$
by matching equations \ref{eqn:rew-lga} and \ref{eqn:quadratic}.
Starting from the need to match
 $\mat{K}_{c,a} = \mat{H}^T\mat{M}$,
we can see that it is not only \emph{always possible} to transform rewards
into goals and outcomes, but it is even under-determined.
There are infinitely many decompositions $\mat{H}^T\mat{M}$ for any matrix $\mat{K}_{c,a}$.
For any choice of $\mat{H}$ and $\mat{M}$, a perfect match $r\!=\!\hat{r}$ can be generated
by the residual terms
\begin{equation*}
 \mat{R}_c = \mat{K}_{c,c} + \mat{H}^T\mat{H} \ \ \ \ \text{and} \ \ \ \  
 \mat{R}_a = \mat{K}_{a,a} + \mat{M}^T\mat{M} \ .
\end{equation*}
For a concrete decomposition of $\mat{K}_{c,a}$ we can consider
its singular value decomposition $\mat{U}\mat{S}\mat{V}^T$ with orthonormal matrices $\mat{U},\mat{V}$
and a positive diagonal matrix $\mat{S}$.
An exemplary decomposition could be to set $\mat{H} = \mat{S}^{\frac{1}{2}}\mat{U}^T$
and $\mat{M} = \mat{S}^{\frac{1}{2}}\mat{V}^T$.
Still, the resulting observation space $\spa{X}$, in which $\hat{f}\!:\spa{A}\!\rightarrow\!\spa{X}$
and $\hat{h}\!:\spa{C}\!\rightarrow\!\spa{X}$ map actions and contexts, is very high-dimensional 
with $\mathrm{min}(p',m')$ dimensions, since $\mat{K}_{c,a}\in\Rm{p'}{m'}$.
However, a dimension reduction is now straightforward based on the SVD: we can 
select the diagonal matrix $\mat{S}'\in{}\Rm{n}{n}$ with the $n$ largest singular values of $\mat{K}_{c,a}$ 
and their respective singular vectors in $\mat{U}'\in{}\Rm{p'}{n}$ and $\mat{V}'\in{}\Rm{m'}{n}$
to approximate $\mat{K}_{c,a} \approx \mat{U}'\mat{S}'\mat{V}'^T = \mat{H}^T\mat{M}$.
$\mat{H}\in{}\Rm{n}{p'}$ and $\mat{M}\in{}\Rm{n}{m'}$ 
can be chosen within the column space of $\mat{U}'$ and $\mat{V}'$ in order to 
project into the $n$-dimensional observation space.
Hence, the latent observation space can be uniquely determined for any number of dimensions $n$,
except for multiple identical singular values.
For sufficiently large $n$, LGA then approximates the reward function arbitrarily well.

\paragraph{Optimal Self- and Goal-Detection}\label{sec:match}
While the selection of the observation space is uniquely determined by the column space of
$\mat{U}'$ and $\mat{V}'$, the positioning of goals and outcomes
\emph{inside} that space, i.e. the precise choice of $\mat{H}$ and $\mat{M}$, is not unique.
We can choose both matrices by means of any transformation matrices $\mat{P}_c, \mat{P}_a \in \Rm{n}{n}$
\begin{equation*}
\mat{H} = \mat{P}_c \cdot{} \mat{U}'^T  , \ \ \ \ \mat{M} = \mat{P}_a \cdot{} \mat{V}'^T  \ ,
\text{ that satisfy}
\end{equation*}
\begin{equation*}
 \mat{P}_c^T \mat{P}_a \!=\! S' \ \ \Rightarrow \ \ 
    \mat{H}^T\mat{M} \!=\! \mat{U}'\mat{P}_c^T \mat{P}_a\mat{V}'^T \!=\! \mat{U}'\mat{S}'\mat{V}'^T \ .
\end{equation*}
We can see our options by representing $\mat{P}_a$ by its singular value decomposition with orthonormal matrices $\mat{u},\mat{v}\in{}\Rm{n}{n}$
and positive diagonal matrix $\mat{s}\in{}\Rm{n}{n}$:
\begin{equation*}
\mat{P}_a = \mat{u}\mat{s}\mat{v}^T \ \ \Rightarrow \ \ \mat{P}_c = \mat{u}\mat{s}^{-1}\mat{v}^T\mat{S}' \ .
\end{equation*}
Using this notation it is easy to show that $\mat{u}$ is irrelevant.
If we insert the above definitions into the reward equation \ref{eqn:rew-lga},
we can see that $\mat{u}$ only appears as $\mat{u}^T\mat{u}$.
Since $\mat{u}$
is orthonormal we get $\mat{u}^T\mat{u}\!=\!\Id{n}$ in any case and can set $\mat{u}\!=\!\Id{n}$
right away. This reflects that a rotation of the entire observation space 
does not change distances in that space, and therefore does not matter for LGA.
It remains to choose an orthonormal matrix $\mat{v}\in\Rm{n}{n}$ and a positive diagonal matrix $\mat{s}\in\Rm{n}{n}$.
$\mat{v}$ and $\mat{s}$ together determine how actions and contexts are precisely projected into the observation
space as goals $\hat{h}(\c)\!=\!\hat{\x}^*$ and action-outcomes $\hat{f}(\a)\!=\!\hat{\x}$. 
Interestingly, $\hat{\x}^*$ and $\hat{\x}$ are thereby scaled ``against'' each other:
Increasing $\mat{s}$ results in scaling \emph{up} $\hat{\x}$, but scaling \emph{down} $\hat{\x}^*$ because
of the appearance of $\mat{s}^{-1}$ in $\mat{H}$.
This operation largely modifies the distance between $\hat{\x}^*$ and $\hat{\x}$, and therefore the value 
of $-||\hat{\x}^*-\hat{\x}||^2$.
Nevertheless this does \emph{not} change the overall reward function $\hat{r}$, 
but shifts parameter ``mass'' between its different terms $-||\hat{h}(\c)-\hat{f}(\a)||^2$, 
$\hat{e}_c(\c)$, and $\hat{e}_a(\a)$. Therefore also the contribution of rewards to $\hat{r}$
by these different terms is changed by the choice of $\mat{v}$ and $\mat{s}$.
Of course, $-||\hat{\x}^*-\hat{\x}||^2$ is the most relevant term in LGA, since it reflects the 
relation between goals and outcomes as main constituents of the control problems. Therefore it should have the most significant 
contribution to $\hat{r}$, while the cost terms $\hat{e}_c(\c)$ and $\hat{e}_a(\a)$ 
should rather be residuals.
Hence, we can choose $\mat{v}$ and $\mat{s}$ such that $\hat{e}_c(\c)$ and $\hat{e}_a(\a)$ 
are minimal, which we can formulate with $l_2$ norms of the matrices $\mat{R}_c$
and $\mat{R}_a$:
\begin{equation*}
 \mat{v}^*, \mat{s}^* = \underset{\mat{v}, \mat{s}}{\operatorname{argmin}}( E_R ) =  \underset{\mat{v}, \mat{s}}{\operatorname{argmin}} \left( \|\mat{R}_c\|_2 + \|\mat{R}_a\|_2 \right)
\end{equation*}
This formulation involves the high-dimensional matrices $\mat{R}_c$ and $\mat{R}_a$,
but can be boiled down to an equivalent lower-dimensional problem.
Inserting the definitions of $\mat{H}$, $\mat{M}$ based on $\mat{v}$ and $\mat{s}$
gives
\begin{eqnarray*}
 \mat{R}_c &=& \mat{K}_{c,c} + \mat{U}'\mat{S}'\mat{v}\mat{s}^{-2}\mat{v}^T\mat{S}'\mat{U}'^T \ \ \text{and} \\
 \mat{R}_a &=& \mat{K}_{a,a} + \mat{V}'\mat{v}\mat{s}^2\mat{v}^T\mat{V}'^T \ .
\end{eqnarray*}
$\mat{v}$ and $\mat{s}$ can only affect $\mat{R}_c$ and $\mat{R}_a$ within the subspace 
spanned by $\mat{U}'$ and $\mat{V}'$. Hence, we can equivalently consider their $n$-dimensional projections 
\begin{eqnarray*}
 \mat{R}_c^{(n)} &=& \mat{U}'^T\mat{K}_{c,c}\mat{U}' + \mat{S}'\mat{v}\mat{s}^{-2}\mat{v}^T\mat{S}' \ \ \text{and} \\  
 \mat{R}_a^{(n)} &=& \mat{V}'^T\mat{K}_{a,a}\mat{V}' + \mat{v}\mat{s}^2\mat{v}^T \ ,
\end{eqnarray*}
and minimize the respective error, such that
\begin{equation*}
 \mat{v}^*, \mat{s}^* = \underset{\mat{v}, \mat{s}}{\operatorname{argmin}}( E_R^{(n)} ) \!=\!  \underset{\mat{v}, \mat{s}}{\operatorname{argmin}} \left( \|\mat{R}_c^{(n)}\|_2 \!+\! \|\mat{R}_a^{(n)}\|_2 \right) .
\end{equation*}
We are currently not aware of a closed form solution to this optimization problem.
It is low-dimensional, though. In practice we found that it can be solved both efficiently
and very effectively by simple gradient descent. Therefore we initialize $\mat{v}\!=\!\mat{s}\!=\!\Id{n}$
and apply gradient descent on $E_R^{_{(n)}}$ with a step width of $0.001$ until numeric convergence.
Thereby only the diagonal values of $\mat{s}$ are considered, and $\mat{v}$ 
is orthonormalized by setting its singular values to $1$ after each update.

\subsection{Algorithm Summary and Interpretation}

Altogether, LGA starts which a universal approximation of the reward or value (i.e. future expected reward) function
function in the quadratic form shown in equation~\ref{eqn:quadratic}.
This mechanism corresponds to value systems supposed to exist in midbrain structures \cite{Schultz1997,daw2006computational},
whereas the reward itself could reflect any extrinsic or intrinsic phenomenon.

The second step is the SVD of $\mat{K}_{c,a}$. Here we select the axis in column and row space 
that have the highest singular values. This corresponds to the axis of inside the action- and
context space that are most significant to the reward/value function. Hence, this step identifies
the low-dimensional observation space in which goals and action outcomes are situated.

The third and last step is to actually locate goals and action outcomes inside the observation space.
Therefore we assemble the matrices  $\mat{H}$ and $\mat{M}$ for goal- and self-detection out of 
the terms $\mat{v}$ and $\mat{s}$. This directly gives the functions $\hat{h}(\c)$ and $\hat{f}(\a)$
and allows to compute $\hat{e}_c(\c)$ and $\hat{e}_a(\a)$ if needed.

We have argued on a purely conceptual level that goals should be considered as abstractions
of reward signals. Latent Goal Analysis provides the computational tool for this idea.
Thereby goals and outcomes are considered latent variables of the reward function,
which allows to view them as compact low-dimensional representations of otherwise high-dimensional actions
and world states. These representations thereby do justice to the semantics of desire and intention
by expressing aspects relevant to reward, and do justice to the achievement semantics
by representing both in a common space in which they are compared.
In the following experiments we will exemplify all of these aspects, and the universality of the approach, 
by showing ($i$) a dimension reduction application based on external rewards in Sec.\ \ref{sec:exp-rec},
and ($ii$) a developmental study with generic and intrinsic information seeking rewards in Sec.\ \ref{sec:exp-sal}
in which highly interpretable representations are learned, and used to bootstrap self-supervised 
motor learning.

\section{Experiment: Dimension Reduction in News Article Recommendation}\label{sec:exp-rec}

Our computational learning formulation shows that goals can be learned as abstractions of rewards.
The algorithm itself is in the first place meant to show the feasibility of this approach.
However, since the formulation is based on spectral decomposition we can immediately apply
it for \emph{dimensionality reduction} of reward-based learning problems.
Our first experiment therefore investigates this ability, for which we take a very \emph{practical}
scenario. Here, the ``task'' is in fact already given by means of extrinsic rewards. We will show that
learned goals can serve as very useful and compact representations of such externally given tasks.
We will contrast this external specification in Sec.\ \ref{sec:exp-sal} where we investigate purely intrinsic 
rewards that to not immediately describe any task.

Our experiment investigates LGA's capability for dimension reduction in a one-step reinforcement learning problem.
This scenario considers a website comprising a certain set $A(t)$ of news articles at each time.
One article can be featured, i.e. \emph{recommended}, at a prominent position on the website.
The task is to select which article's teaser (action $\a\!\in{}\!A(t)$) should be on the featured position.
A \emph{recommender system} is supposed to select these actions such that 
the probability that the website visitor interacts with it (e.g. clicks on the teaser, or purchases a good) is maximized,
such that the earning of the operating web company rise in the end. 
In order to perform such selection specific to the visitor, there is in many cases
information like country, age, or previous click-history (context $\c$) available due to IP-address, cookies,
and a login at the website.
With such information a reward function $r(\c,\a)$ can be estimated that resembles 
the click probability. 
Dimensionality reduction, however, is crucial in this domain: both context and action are typically
very high-dimensional, but any recommender system must react extremely quickly to thousands or millions of visits of a webpage.
This can only be achieved if the dimension of $\c$ and $\a$ is reduced to allow for an efficient 
evaluation of $r(\c,\a)$.

Dimension reduction in reinforcement learning is a tedious issue.
One attempt has been to learn reduced rank regression of transition probabilities to guide
exploration \cite{nouri2010dimension}, but which can not directly consider the features' \emph{relevance} to reward achievement.
Purely unsupervised schemes like PCA, or slow-feature analysis \cite{Legenstein2010}
are frequently applied for state-space dimension reduction, but also cannot account for the actual reward-relevance. 
Reward-modulated versions of such learning rules \cite{bar2003information} can account 
for the reward-relevance at least to some extent, but are limited to simple correlations.
None of these methods can effectively reduce the dimension of \emph{actions} because actions
do not have a naturally observable probability distribution (except when expert demonstrations are given \cite{bitzer2011nonlinear}).
The only models that can consider states, actions, and reward at the same time estimate
the reward function based on bi-linear regression and reduce the rank of the parameter matrix \cite{Koren2009,Chu2009-Bilinear,Chu2009-CaseStudy}.
From a perspective of dimension reduction only, these models are similar to our approach, although coming from an entirely different direction. 
We will show experimentally that our method yields significantly more compact representations in
a practical scenario \cite{WebscopeR6B}.

\subsection{Material and Method}

For this experiment we use the 
``Yahoo!\ Front Page Today Module User Click Log Dataset, version 2.0'' \cite{WebscopeR6B},
which comprises recordings of the click behavior on \emph{yahoo.com}'s front 
page from 15 consecutive days in October 2011, from which we utilize the first day 
of recordings only.
This recording contains a total number of $T\!=\!1.607.525$ events. Each
event contains the actually displayed news-teaser, the set of currently available news,
a set of visitor features, and the visitor's decision to click on the teaser ($r\!=\!1$) or not ($r\!=\!0$). 
49 different teasers have been shown in this period, which we represent 
as $m\!=\!49$ dimensional actions $\a$ encoded with a ``1-of-$m$'' scheme.
The events contain a total number of 116 binary features about the visitor,
which are fully \emph{anonymized} in the data, i.e. their individual meanings are not revealed by Yahoo. 
Using this data, we estimate $\mat{K}$ using batch-gradient descent on the empirical 
error $E[ (\hat{r}(\c,\a,\mat{K}) - r)^2 ]$. We applied 10000 epochs of training 
with a gradient step width 0.01 starting from zero initial parameters. After that 
the parameters were fine-tuned by applying a whitening on the contexts and continuing
batch regression for another 1000 epochs with step width 0.0001.

As a baseline, we applied a bi-linear regression model $\hat{r}(\c,\a)\!=\!\psi_c(\c)^T\!\cdot{}\!\mat{B}\!\cdot{}\!\psi_a(\a)$
that was trained with the same procedure as the quadratic model.
Such bi-linear regression (BLR) models have previously \cite{Koren2009,Chu2009-Bilinear} been used to 
reduce the dimension in recommender scenarios: the matrix $\mat{B}$ can
be decomposed into $\mat{U}_B\mat{S}_B\mat{V}_B^T$ by singular value decomposition, after which only 
the $n$ most significant dimensions are kept. This decomposition allows to rewrite 
$\hat{r}$ as
\begin{equation*}
 \hat{r}(\c,\a) = \left(\mat{S}_B^{\frac{1}{2}}\mat{U}_B^T \cdot{} \psi_c(\c)\right)^T \cdot{} \left(\mat{S}_B^{\frac{1}{2}}\mat{V}_B^T \cdot{} \psi_a(\a)\right) 
\end{equation*}
Very similar to the LGA approach,  this method can be interpreted as projecting contexts
and actions into a common space, which has been referred to ``partworth'' space \cite{Chu2009-CaseStudy}. 
In this space, the comparison is done by a scalar product, whereas LGA measures a direct distance. 
The ``partworth'' space therefore does not encode \emph{goals} because 
they cannot be definitely achieved
(the scalar product is not bounded). 
The evaluation cost for both
models is the same: both involve matrix-vector multiplications of the same size;
the square-distance in LGA is only exactly one floating point operation more expensive than 
the scalar product. 
As further baselines we used PCA to reduce the dimension 
of the context-space before applying either quadratic or bi-linear regression. In the PCA condition
the dimension of actions cannot be reduced.

\begin{figure}
\begin{center}
\includegraphics[width=0.95\columnwidth,trim = 1.5cm 1.6cm 1cm 2.3cm, clip=true]{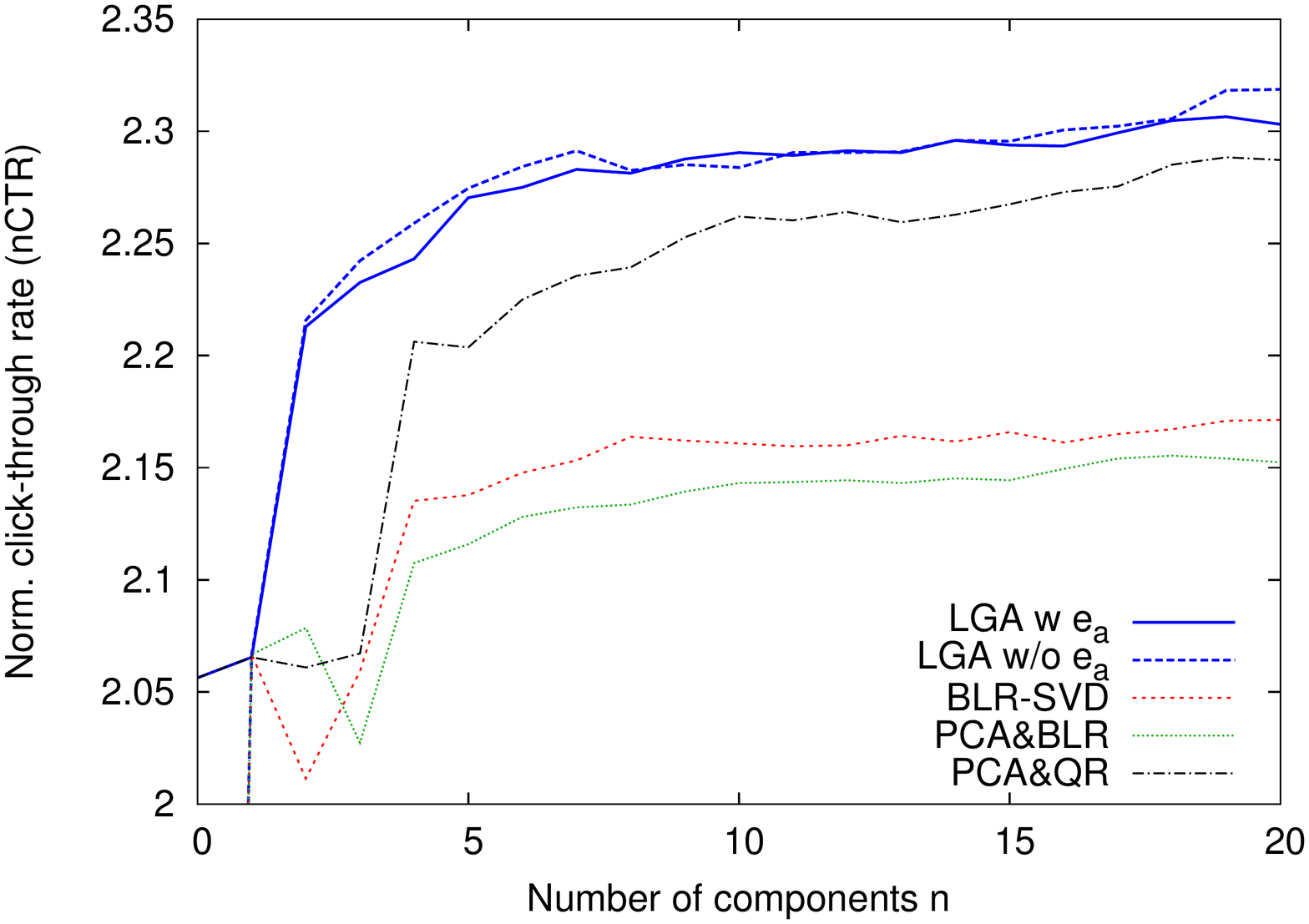}
\caption{Results for the Yahoo! Webscope R6B data set. 
	Estimated normalized click-through rates for LGA, BLR decomposition, and PCA depending on the number of selected components.}
\label{webscope}
\end{center}
\end{figure} 

\subsection{Results}

In order to evaluate the effectiveness of each technique we need a domain specific evaluation metric,
i.e.\ how many visitors' clicks both approaches can generate.
For each method we can denote the policy to choose a news-teaser $\a$ based on the user information $\c$ as $\pi(\c)=\operatorname{argmax}_{\a\in{}A(t)}\hat{r}(\c,\a)$, where $A(t)$ is the set of articles 
available at the time $t$ of the page visit.
A natural performance metric then is the \emph{click-through rate} 
$\operatorname{CTR}_\pi\!=\!N_\pi^+/N$, where $N$ is the total number of page visits
and $N_\pi^+$ is the number of clicks generated by selecting teasers with $\pi$.
Yet, this measure can only be thoroughly measured when the policy is run \emph{online}
on the webpage.
For an \emph{offline evaluation} \cite{Chu2009-CaseStudy} we can estimate
the performance by counting how often an actually clicked teaser would
have been also recommended by the policy:
\begin{equation*}
 \operatorname{nCTR} = \frac{ \operatorname{CTR}_\pi }{ \operatorname{CTR}_{\%} } 
\approx{} \frac{ |\{r_t=1\wedge{}\a_t=\pi(\c_t)\}| }{ \sum_t(r_t\cdot{}|A(t)|^{-1}) }, 
\end{equation*}
which is baselined against the performance $\operatorname{CTR}_{\%}$ of a uniform random 
strategy.
Fig.\ \ref{webscope} shows that LGA achieves a substantially better performance
than the bi-linear model decomposition (BLR-SVD). In fact, already $n\!=\!0$ components for LGA, which corresponds
to only evaluating $\hat{e}_a(\a)$, suffice to reach a factor of $2.06$ compared
to chance level. This reflects the simple fact that some articles are more interesting than others.
LGA then quickly improves to $\operatorname{nCTR}\!>\!2.25$ for $n\!\geq\!5$ components
and further improves to $\operatorname{nCTR}\!>\!2.3$.
The bi-linear model  requires $n\!\geq\!8$ components to reach only $\operatorname{nCTR}\!>\!2.15$
with only minimal further improvement for more components.
It might be considered unfair to allow the $\hat{e}_a(\a)$ term for LGA, whereas 
the bi-linear models do neither comprise nor require such an additional measure.
In the recommender scenario, however, it is reasonable because $\hat{e}_a(\a)$ only 
needs to be computed \emph{once} when an article is published. 
The cost for evaluating the policy on a page visit is the same, since 
both LGA and BLR need to multiply the visitor context with a $(n\times{}p)$
matrix. LGA then computes a distance in $n$ dimension and BLR a scalar product in $n$
dimensions, which also has the same cost. However, in this particular setup LGA
achieves equally high performance (for $n\!>\!0$) when the cost term is omitted (see Fig.\ \ref{webscope}).
Both LGA and bi-linear decomposition outperform their counterparts with unsupervised PCA
on the states before running bi-linear (BLR) or quadratic (QR) regression. Interestingly,
PCA\&QR substantially outperforms the standard BLR decomposition approach, which shows the
high expressiveness of quadratic regression in general.  
PCA\&QR has, however, an increased cost of evaluating $\hat{r}(\c,\a)$ for a new context: $O(n^2\!+\!n\!\cdot{}\!p)$ instead of $O(n\!+\!n\!\cdot{}\!p)$
for all other methods.
LGA still extracts more compact representations from the quadratic regressor and outperforms
PCA\&QR.

\subsection{Discussion}

We can conclude that LGA allows for an effective dimensionality reduction in 
the online news recommender setup, in which it outperforms the standard bi-linear model
in terms of generated clicks. The margin is thereby numerically not very high in the range of $5$-$10\%$,
but which is still highly significant to the domain since clicks are directly related to a website's monetary income.
Other studies on recommender systems have reported much higher absolute values of $\operatorname{CTR}$ for other benchmarks, which
suggests that the data set used here is rather hard. A possible reason is that there are no features for the actions,
but only identities. Features to relate similar articles could therefore further increase 
both the absolute $\operatorname{CTR}$ as well as the margin between different methods.

Goals seem to be useful abstractions in this setup. Within our framework goals and action outcomes
serve as low-dimensional, compact representations of what defines the reward. In this particular case,
though, it is impossible to interpret the goals' semantics, simply for the reason that we do not
know the meaning of the context features and the content of the articles, all of which had been previously
anonymized. If we \emph{would} know them, however, we could interpret the learned observation space
as a space of different \emph{topics of interest} of users. From the webserver's perspective the user (its context $\c$)
comes with a topic of interest that is described as \emph{goal} 
in this space. The webserver has to-be-shown articles as actions which are described in the topic space
by means of action outcomes: the ``result'' of choosing a particular article is to hit a certain point
in the interest space. The achievement semantics describe that if both match, the user is \emph{interested} 
and likely to generate a reward (click) to the system.

\begin{figure}
\begin{center}
\captionsetup[subfigure]{margin=0.5pt,format=hang,parskip=0.5pt,singlelinecheck=true}
\subfloat[][Arm image, saliency, smoothed saliency]{\label{scheme-lga-gb-saliency}
\includegraphics[width=0.6\columnwidth]{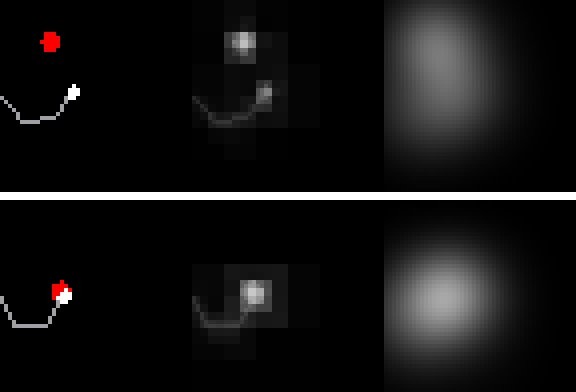}
} \vspace*{-0.2cm}\\
\subfloat[][Schematic organization of the experiment]{\label{sf:arm-setup}
\includegraphics[width=0.7\columnwidth,trim = 0cm 0.5cm 0cm 0.6cm, clip=true]{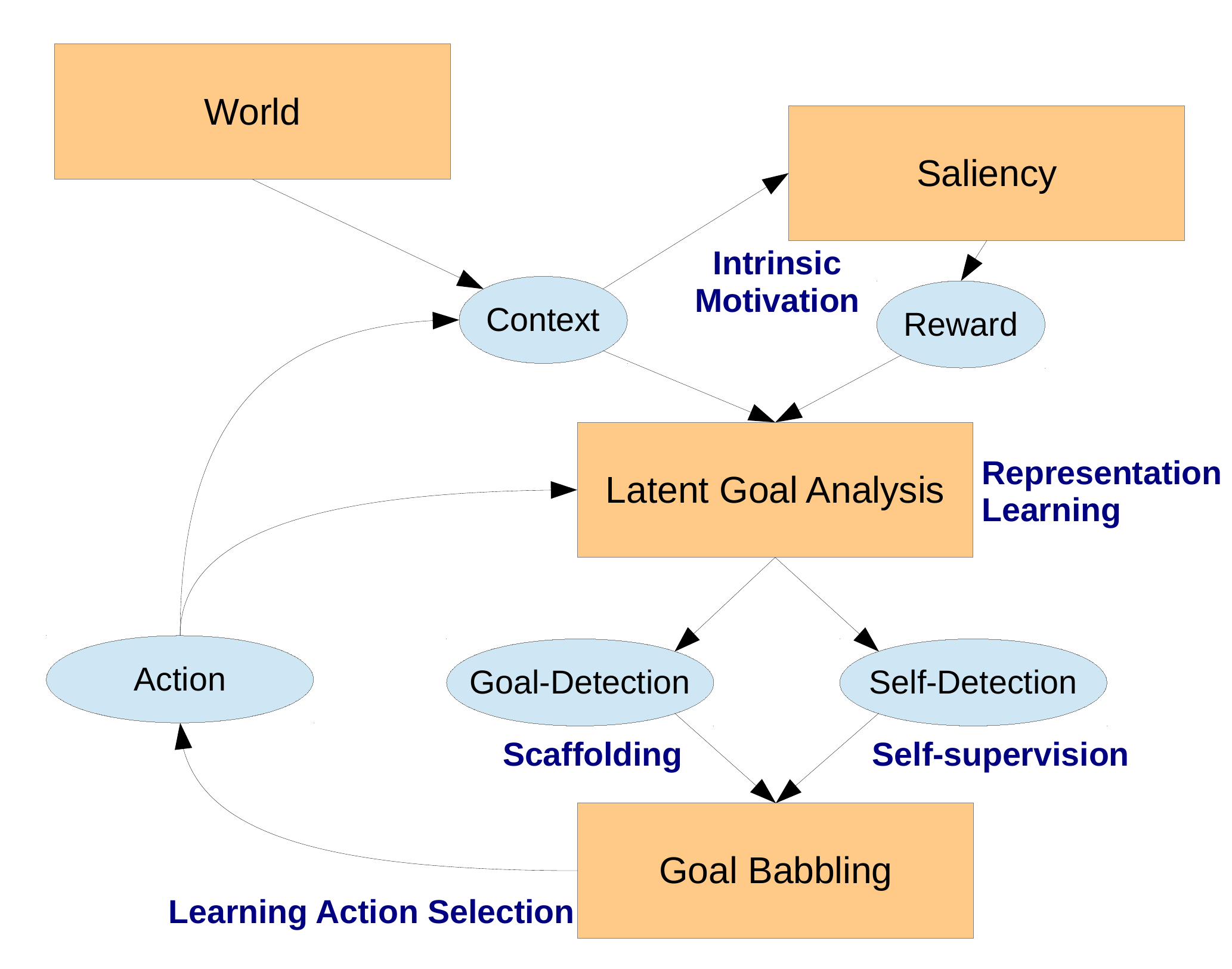}\vspace*{-0.2cm}
}
\caption{We consider saliency to generate an information seeking reward, from which
	LGA learns representations that are used by Goal Babbling.}
\label{sf:sal-setup}
\end{center}
\end{figure} 

\section{Experiment: Goal Development in Reaching}\label{sec:exp-sal}

\begin{figure*}
\begin{center}
\captionsetup[subfigure]{margin=0.5pt,format=hang,parskip=0.5pt,singlelinecheck=true}
\subfloat[][During learning, the system develops from initially no coordination (left) to 1d reaching (middle) to 2d reaching (right).]{\label{arm-examples}
\includegraphics[width=0.3\textwidth,trim = 0cm 3.5cm 0cm 0cm, clip=true]{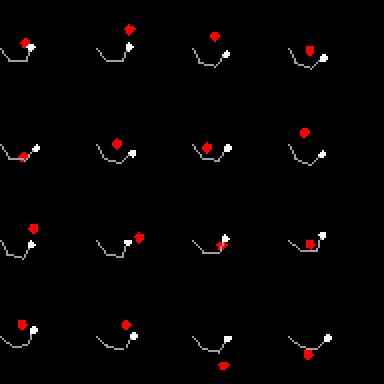}
\includegraphics[width=0.3\textwidth,trim = 0cm 3.5cm 0cm 0cm, clip=true]{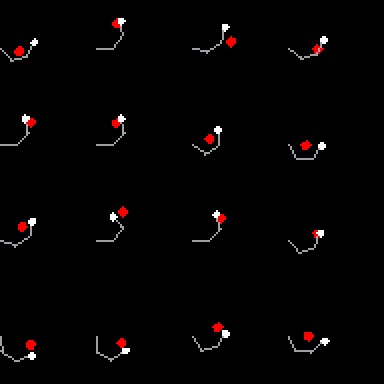}
\includegraphics[width=0.3\textwidth,trim = 0cm 3.5cm 0cm 0cm, clip=true]{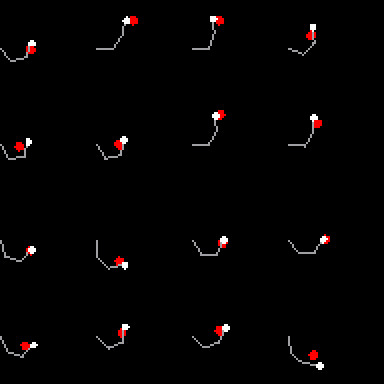}
} \vspace*{-0.2cm}\\
\subfloat[][The self-detection encodes the hand position.]{\label{self}
\includegraphics[width=0.32\textwidth,trim = 1.5cm 1.5cm 1cm 2.3cm, clip=true]{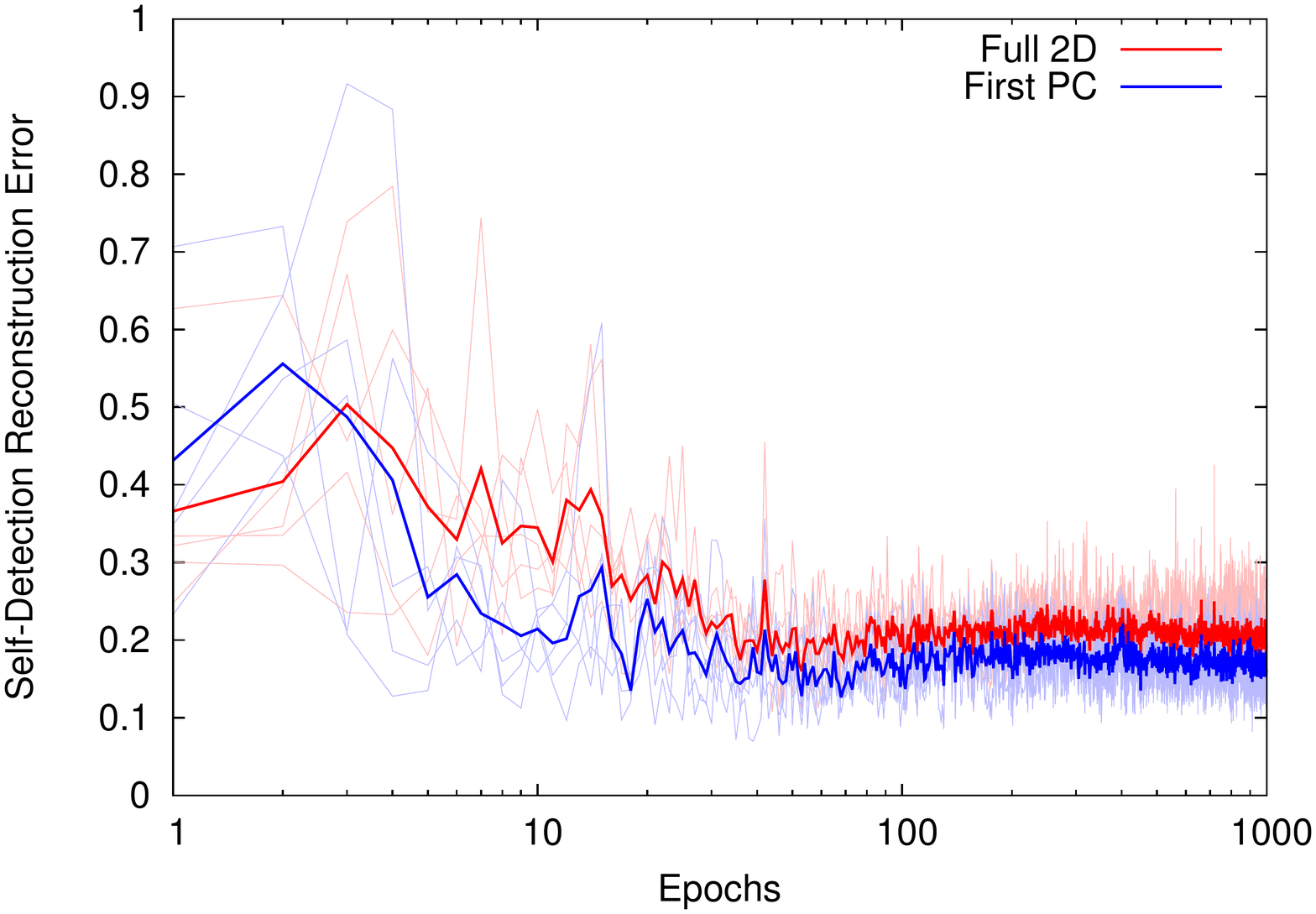}
} 
\subfloat[][The goal-detection encodes the object position.]{\label{goal}
\includegraphics[width=0.32\textwidth,trim = 1.5cm 1.5cm 1cm 2.3cm, clip=true]{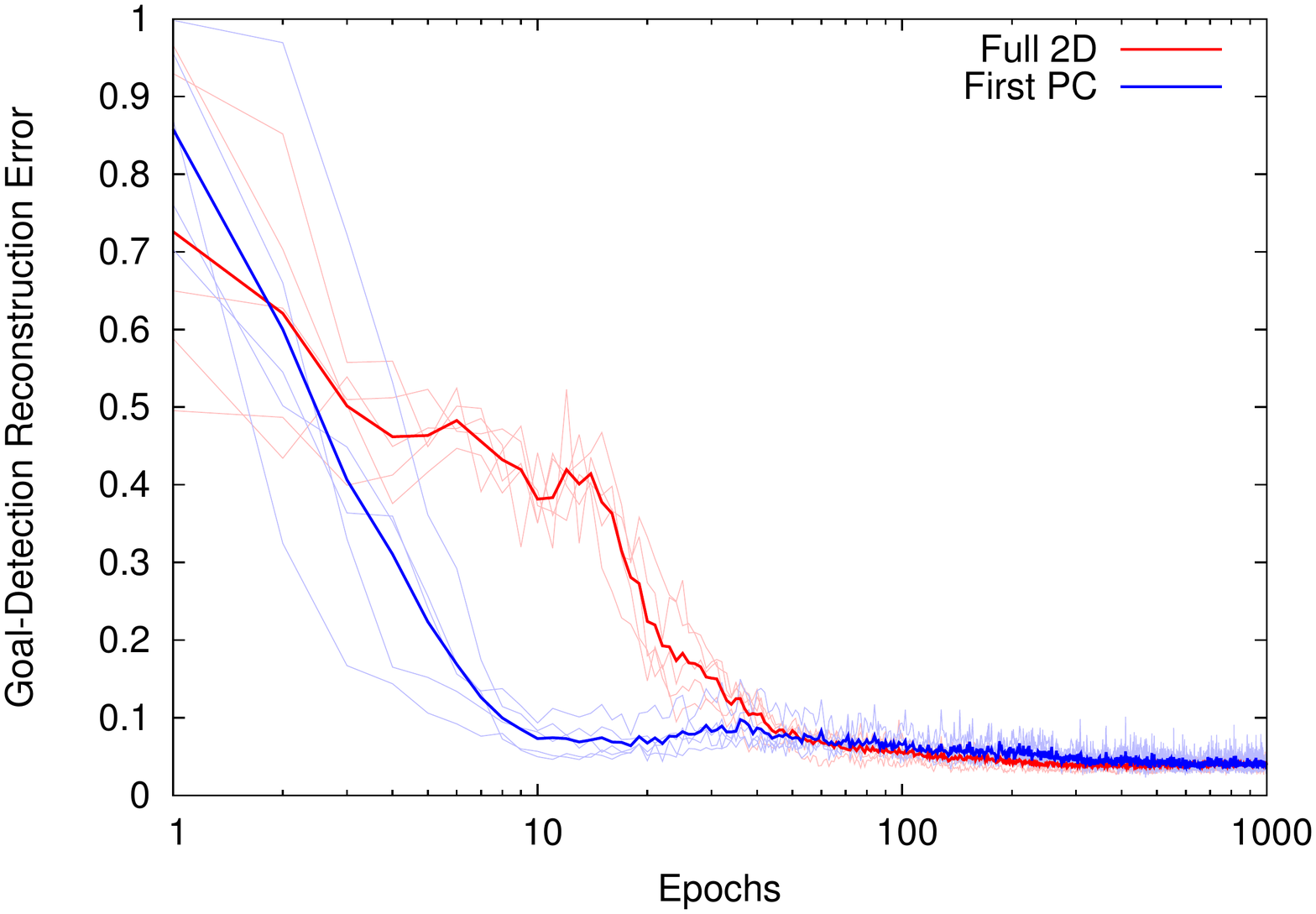}
}
\subfloat[][Hand-object contact rate via the inverse model.]{\label{reach-rate}
\includegraphics[width=0.32\textwidth,trim = 1.5cm 1.5cm 1cm 2.3cm, clip=true]{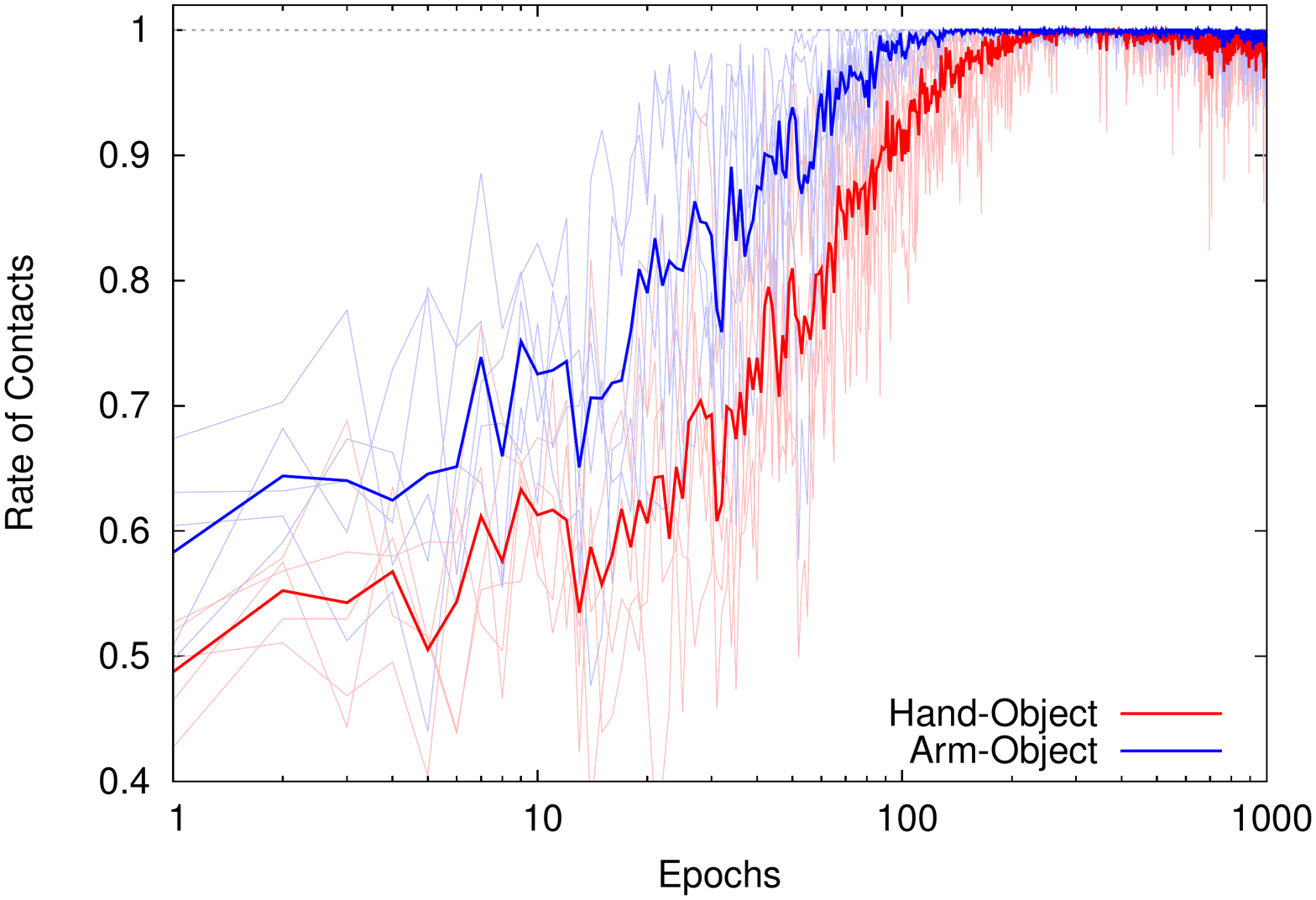}
}
\caption{Goal- and self-detection learn to represent the object and hand position respectively. The first principal component 
	in the internal coordinate system encodes the top-down coordinate.
	Goal babbling uses these abstractions to learn an
	  inverse model that consistently makes contact with the object.}
\label{sal-results}
\end{center}
\end{figure*}

In the last section we have shown that LGA can extract goals as useful abstractions
from extrinsic rewards which already specify a concrete task. 
But how could an agent learn goals when the task is not that explicitly or not at all 
given already in the beginning?
In this section we contrast the previous experiment with \emph{intrinsic rewards}
that do not describe any particular task.
We consider
visual saliency as a reward for a simple robot arm to implement \emph{information seeking} behavior \cite{Gottlieb2012} and
show that it leads to meaningful goal- and self-representations.
Saliency measures have already been shown to permit a self-detection of the own 
end-effector \cite{HikataAsada2008}, simply because looking at the own hand is ``interesting''.
Here we extend this finding by considering an object at the same time. It turns out that more
interesting than looking at the hand \emph{or} the object is to look at \emph{both}
closely together (compare Fig. \ref{arm-examples}\subref{sf:arm-setup} top and bottom).
We show that LGA thereby develops a detection of an external object as goal, and
a self-detection of the own hand. These representations are thereby already \emph{utilized} 
by means of goal babbling \cite{Rolf2010-TAMD} in a closed loop, which
results in the \emph{emergence of goal-directed reaching}.

\subsection{Setup}

The basic scenario is shown in Fig. \ref{sf:sal-setup}.
We consider a simple robot arm with three joints (segment length $1/3$ each), such that \emph{actions} are 
the joint angles $\a_t\!\in\!\Rv{3}$. We refer to the effector's actual position (that is at no time
explicitly known the learner) in cartesian coordinates as $x_t\!\in\!\Rv{2}$.
A salient object is placed somewhere in the scene at coordinates $o_t\!\in\!\Rv{2}$.
Arm and object together are rendered into a 48x48 pixel image.
Generically we could think of this image as \emph{context} in terms of visual perception.
However, considering raw 2300 dimensional visual input for learning is neither computationally 
feasible nor very biologically plausible.
For this experiment we assume a certain extent of image processing that has already 
identified the object and hand coordinates as keypoints in this image.
The context for learning comprises these basic coordinates plus additional noise dimensions 
to challenge learning.
At every timestep $t$ the agent is assumed to be still in position $x_{t-1}$, with the object 
at position $o_t$. 
Hence, the learner's context is $\c_t\!=\!(o_t; x_{t-1}; \varepsilon) \!\in\! \Rv{6}$
with gaussian noise $\varepsilon\!\in\!\Rv{2}, \varepsilon_i\!\sim{}\!\mathcal{N}(0.5, 0)$.
For the to be estimated functions $\hat{h}$, $\hat{f}$,  $\hat{e}_c$, and $\hat{e}_a$ we
use a locally-linear learning formulation identical to \cite{Rolf2011-ICDL} with receptive field 
radii $2.0$, $0.25$, $0.5$, and $0.5$ respectively.
As a design choice we selected $n\!=\!2$ components to be extracted from the $6$ dimensional 
context and $3$ dimensional action.

\paragraph*{Reward}

The learner's reward $r_t$ is computed using a simple saliency model based 
on difference-of-gaussians (simplified from Itti's model \cite{Itti98}). 
After the agent has received the context $c_t$ (containing an image of the old action $a_{t-1}$) and 
selected a new action $a_t$, we compute a reward based on the ``after-action'' image containing the object
and the new action $a_t$.
Then, we compute a pyramid of gaussians: The image is smoothed with a 5x5 gaussian kernel and scaled down by a factor of 2. This procedure is repeated 4 times.
The saliency map (Fig. \ref{sf:sal-setup}\subref{scheme-lga-gb-saliency}, middle) is computed out of these 
5 images (original \& 4x smoothed) by taking the the difference between any two of them, and adding 
up the amplitudes of those differences.
Considering the most salient point would mean to look at the most interesting pixel.
However, we assume that the agent does not just attend to a single visual receptor but 
rather a \emph{region} in the visual scene. Therefore we smooth the saliency map on a large 
scale (Fig. \ref{sf:sal-setup}\subref{scheme-lga-gb-saliency}, right) with a gaussian filter with $\sigma\!=\!10$ pixels
width which models the total width of the agent's visual field. The \emph{highest value} of this \emph{smoothed saliency} 
therefore measures how much information the agent can have in its visual field.
We manually normalized the scale of these values such that they approximately lie in a range $[0;1]$ and consider these
the rewards $r_t$.

\paragraph*{Online Learning Algorithm}

For this experiment we developed an \emph{online} algorithm for LGA. 
While the previously introduced algorithm is useful to show the theoretic 
feasibility of LGA and for batch processing, it is not suited for the 
closed loop processing intended in this experiment.
Here we use a simple gradient descent algorithm to estimate 
the goal- and self-detection \emph{directly} (as opposed to learning a full reward $r(\c,\a)$ model first).
We consider an agent that observes 
samples along time $t$. The agent perceives a context $\c_t$, 
executes an action $\a_t$, and receives a reward $r_t\!=\!r(\c_t,\a_t)$ based 
on a hidden reward function $r(\c,\a)$.
The agent is supposed to learn the functions $\hat{h}$, $\hat{f}$,  $\hat{e}_c$, and $\hat{e}_a$
such that the observed reward $r_t$ is explained by them according to Eqn. \ref{eqn:rew-transform-est}.
This can be done by reducing the \emph{reward-prediction error}:
\begin{equation*}
 E_t(r_t, \c_t,\a_t) = || e_t(r_t, \c_t,\a_t) ||^2 =  || r_t - \hat{r}(\c_t,\a_t) ||^2 \ \ . \label{eqn:rew-error}
\end{equation*}
We denote the learnable parameters of $\hat{h}$, $\hat{f}$,  $\hat{e}_c$, and $\hat{e}_a$
as $\theta_h$, $\theta_f$, $\theta_c$ and $\theta_a$ respectively.
For an initial symmetry-breaking (due to the invariance of internal rotation and translation)
it is necessary to initialize $\theta_h$ and $\theta_f$ with small random values.
From this point on, simple gradient descent on $E$ can succeed to estimate the functions.
However, we need to consider that the values of $\hat{e}_c$ and $\hat{e}_a$
have to be kept small. For this purpose we use a simple decay term similar to weight-decay
often used in neural networks: parameter values are not only adapted by the 
error-reduction signal, but also by a decay of some $\epsilon\!\in\![0;1)$ portion of their value.
Since any reward mass that decays from $\hat{e}_c$ and $\hat{e}_a$ needs to be explained 
by $|| \hat{h} \!-\! \hat{f} ||$ instead, we \emph{add} (reverse sign) the decay values to 
the learning signals of $\hat{h}$ and $\hat{f}$.
The resulting gradient rule with rates $\eta_d$ and $\eta_c$ is:
\begin{eqnarray*}
 \Delta\theta_f & \!=\! & + \eta_d \cdot \big(e_t \!+\! \epsilon\!\cdot{}\!\left[\hat{e}_a(\a_t)\!+\!\hat{e}_c(\c_t)\right] \big) 
			\!\cdot{}\! (\hat{\x}^*_t \!-\! \hat{\x}_t) \!\cdot{}\! \frac{\partial \hat{f}(\a_t)}{\partial \theta_f} \\
 \Delta\theta_h & \!=\! & - \eta_d \cdot \big(e_t \!+\! \epsilon\!\cdot{}\!\left[\hat{e}_a(\a_t)\!+\!\hat{e}_c(\c_t)\right] \big) 
			\!\cdot{}\! (\hat{\x}^*_t \!-\! \hat{\x}_t) \!\cdot{}\! \frac{\partial \hat{h}(\c_t)}{\partial \theta_h} \\
 \Delta\theta_a & \!=\! &   \eta_c \cdot (e_t - \epsilon\cdot{}\hat{e}_a(\a_t)) \cdot \frac{\partial \hat{e}_a(\a_t)}{\partial \theta_a} \\
 \Delta\theta_c & \!=\! &   \eta_c \cdot (e_t - \epsilon\cdot{}\hat{e}_c(\c_t)) \cdot \frac{\partial \hat{e}_c(\c_t)}{\partial \theta_c} \ ,
\end{eqnarray*}
in which the last term in each formula is depending on the function approximation method.
If the decay term is disabled ($\epsilon\!=\!0$), these formulas correspond to ordinary gradient descent on $E$.
The decay term balances the contribution of all terms such that 
goal- and self-detection take the dominating role in explaining the reward. The term 
$-|| \hat{h} \!-\! \hat{f} ||^2$ can, however, not model arbitrary reward functions alone.
In particular, this negative distance can only account for \emph{numerically} negative rewards.
Modeling numerically positive rewards requires the terms $\hat{e}_c$ and $\hat{e}_a$ to 
shift the entire estimate $\hat{r}$ by a constant. If the decay term is used this process 
can never fully reach the necessary shift. In order to still permit a reasonable learning signal for $\hat{h}$ and $\hat{f}$,
we introduce a new and purely scalar term $k$ into the reward estimation. 
This term is not effected
by the decay, but can shift the reward estimate such that $-|| \hat{h} \!-\! \hat{f} ||^2$ can be used 
to model the shape of the reward function:
\begin{eqnarray*}
\hat{r}(\c,\a) & = & -|| \hat{h}(\c) - \hat{f}(\a) ||^2 + \hat{e}_c(\c) + \hat{e}_a(\a) + k  \\
 \Delta k &=&   \eta_c \cdot e_t
\end{eqnarray*}
In this experiment we utilize only $\hat{h}$ and $\hat{f}$, whereas $\hat{e}_a$ 
could potentially be used to select cost-optimal actions for the same goal.
The term $\hat{e}_c$ (and $k$) is not directly useful, but needs to accompany the estimation
when approximating any possible reward function as shown in Sec.\ \ref{sec:lga}.

\paragraph*{Procedure}

We conducted this experiment with 5 independent trials with $t\!=\!10^6$ samples each.
During a continuous movement of the object in the visual scene we performed 
continuous online learning of the LGA with learning rates $\eta_d\!=\!0.015$,
$\eta_c\!=\!0.005$, and $\epsilon\!=\!0.05$. While the entire procedure is possible in an online
fashion only, we decided to perform an additional consolidation phase to speed up learning.
Therefore the generation of new samples is interrupted every epoch of 1.000 samples, and the last 10.000
samples are presented in random ordering 10 more times.

LGA describes how an agent can learn internal representations of
goals and self. It does \emph{not} instantaneously describe a strategy to select actions.
However, we can use the goal- and self-detection to perform self-supervised motor learning:
the executed actions $\a_t$ and estimated outcomes $\hat{\x}_t=\hat{f}(\a_t)$ allow 
to generate a supervised learning signal for common methods of motor learning \cite{Nguyen2011},
that can be used together with the estimated goal $\hat{\x}^*_t=\hat{h}(\c_t)$ to
perform goal-directed motor-control.
Here we utilize a previous algorithm for \emph{goal babbling} \cite{Rolf2011-ICDL},
that also utilizes the goals in order to scaffold learning.
This algorithm learns an \emph{inverse model} $g\!\!:\!\hat{\x}^*\!\mapsto\!\a$ from self-generated examples $(\hat{\x}_t,\a_t)$.
The actions are selected by trying to accomplish the goal by means of the inverse model
plus some exploratory noise $N_t$: $\a_t \!=\! g(\hat{\x}^*_t) \!+\! N_t(\hat{\x}^*_t)$.
For this we use a learning rate 0.02, local model distances 0.15 and exploratory
noise with amplitude 0.15 (see~\cite{Rolf2011-ICDL}).

The entire organization of the experiment is shown in Fig.~\ref{sf:sal-setup}\subref{sf:arm-setup}.
The world provides an object that gets encoded in the context together with the agent's last
action. The agent's saliency system generates an intrinsic reward for information seeking. 
Latent Goal Analysis extracts the reward-relevant 
information from the action (self-detection) and disentangles the goal from other information in
the context in order to explain the reward by the relation between goal and self.
Self-detection and estimated goals are then used for 
self-supervised goal babbling in a closed loop.

\subsection{Results}

We ran an evaluation of every 1.000 samples between two consolidation steps. We investigated three questions:
($i$) What does the self-detection encode?
($ii$) What does the goal-detection encode?
($iii$) What behavior results from that abstractions?
In order to investigate ($i$) and ($ii$) we checked how well 
the \emph{internal} representations of outcomes $\hat{\x}$ and goals $\hat{\x}^*$
describe values of the \emph{actual} effector position $\x$ and object position $o$.
Even if the internal variables encode them perfectly there can be arbitrary shifts and translations in the 
internal coordinate system.
Therefore we computed the best linear fit $L$ from internal representations to actual variables.
We assessed the quality of the encoding by the normalized root-mean-square error (NRMSE) $\sqrt{E\left[||L_x(\hat{\x})-\x||^2\right]}/\sqrt{Var\left[\x\right]}$ (correspondingly for $\hat{\x}^*$ and $o$).
If this error is $0$, the value of the actual variable can be perfectly (linearly) predicted from 
the internal one: the internal representation encodes the actual variable. A value of $1.0$ 
means that the prediction gives an error in the range of the variable's variance, which indicates 
that the internal variable does not encode the actual one at all.

Results for the self-detection are shown in Fig. \ref{self}. Learning a 
representation of the robot's own hand requires a strongly non-linear multi-dimensional mapping. 
Results show that already in the very beginning there seems to be a certain extent 
of encoding with errors significantly below $1.0$. However, this results merely from the 
low versatility of actions in the beginning. Goal babbling initially chooses actions close 
to a single posture since it is not sufficiently trained yet. The outcomes of such locally distributed postures 
can to a limited extent be predicted with the randomly initialized self-detection.
After approx. 50 epochs the values stabilize around $0.2$ which means that $80\%$ of the actual effector-position's
variance can be explained by the internal representations. At later stages there is a minimal increase
of the error values which is because goal babbling learns to use more and more different and wide-spread 
postures. Hence, the population gets less local and harder to describe due to non-linearities.
Latent Goal Analysis after all succeeds to learn the robot arm's forward function from
joint angles to effector position by just using the saliency reward.
We additionally investigated the encoding by checking what different coordinate axes in the 
learned representation encode. The blue line in Fig. \ref{self} shows the prediction of the 
effector's top/down coordinate from just the highest variance principle component of the internal 
representation. Low errors indicate that this axis indeed encodes top/down movements.

If LGA should learn a goal representation that describes goal-directed
reaching, then the extracted goals $\hat{x}^*$ should encode the object position $o$. This could seem simple because the object location 
is already directly encoded in the context $\c$. However, this variable still needs 
to ($i$) be identified as the relevant one among other entries (noise and previous hand-position) in the context, 
and ($ii$) set into the right relation (i.e. orientation, shift, scaling) to the self-detection. 
In particular at later stages of learning in our experiment the own hand-position strongly correlates 
with the object position due to goal-directed reaching, so that keeping track of the 
right variable is far from trivial.
Results in Fig. \ref{goal} show that at the time of initialization the object position is not at all encoded
in the goal-detection.
Then, the strongest principal component 
of $\hat{x}^*$ quickly coincides with the top-down axis of the object position (blue lines).
Few epochs later, the goals' 2D values (red lines) do largely encode the actual object position with errors around $0.05$.

Results so far show that the robot's hand position and the object position are indeed 
found as representations for self and goals. In order to check 
how well they fit together (i.e. whether they are in the right geometric relation inside
the observation space), we checked the behavior generated by goal babbling as a result of both abstractions.
In order to perform an analysis that excludes exploratory noise (to check the representations themselves)
we evaluate the combination of goal-detection $\hat{h}$ and inverse model $g$ (learned based on $\hat{f}$). The function $g(\hat{h}(\c))$ suggests actions $\a$ for any context $\c$. Hence we can check those actions 
and see whether they correspond to a reaching act towards the object position encoded in $\c$.
We counted for the contexts within each epoch how often the actions led to a contact 
of either hand and object, or the whole arm and the object (based on the geometries and 
sizes in Fig. \ref{sf:sal-setup}\subref{scheme-lga-gb-saliency}).
Results in Fig. \ref{reach-rate} show that learning rapidly seeks for contacts of arm and object first,
and shortly later establishes a 100\% contact rate of the robot's hand and the object.
Example performances shown in Fig.\ \ref{arm-examples} thereby give an explanation for the 
not perfectly accurate relation between self-detection and actual effector position despite 100\% contact rate:
There is not need to encode the ``tool-center'' position of the hand -- as it is usually done
in robotics due to the lack of more adaptive representations. Rather, our model learns to contact objects
close to the shoulder with the inner part of the effector, while far away objects are contacted with the outer
part of the effector.

\subsection{Discussion}

Latent Goal Analysis together with goal babbling indeed produces representations as well as 
inverse models that correspond to goal directed reaching in this experiment. 
Remarkably, all of this is bootstrapped 
from a task-unspecific reward based on visual 
saliency as a sole original learning signal. Abstractions bootstrapped by LGA 
are then used as \emph{self-supervised} learning signal for goal babbling.
This experiment therefore has a dual finding: Firstly, we show that saliency as
an information seeking reward results in goal-directed reaching \emph{behavior}
as seen from a \emph{distal perspective}. While it had been reported that saliency 
can account for self-detection \cite{HikataAsada2008} of the own hand, we are not 
aware of studies also showing goal-directed behavior as a direct consequence.
Secondly, we find that this information seeking reward \emph{together} with Latent Goal
Analysis can account for the \emph{emergence of internal self and goal representations} (proximal perspective)
as they are typically preprogrammed in robotics, and pre-supposed in computational neuroscience of internal
models.

\section{Discussion}

The autonomous development of goals is a fundamental issue in developmental robotics.
This paper has proposed a detailed conceptual framework and a mathematical operationalization
for agents to learn goals themselves.
Our main argument therefore is to consider goals as high level \emph{abstractions}
of lower level mechanisms of intention such as reward and value systems which themselves
could origin from either external rewards or intrinsic motivation measures.
We emphasized the need to consider and learn goals alongside with self-detection 
of the own actions' outcomes. Both can then be compared in a common space.
We suggest that goals and outcomes together can be learned by considering them as latent 
variables (i.e. abstractions) that can explain an observed or expected future intrinsic or external reward.
Our computational approach Latent Goal Analysis operationalizes this thought. We have therefore
shown the universal existence of the transformation from rewards to goals.

We have shown two very different experiments with this learning formulation: On the one hand
a concrete task that is already specified by external rewards, and for which LGA can learn 
goals as very compact and practically useful representations.
On the other hand we have shown that considering mere visual saliency as a generic,
task-unspecific information seeking reward to be processed by our framework
leads to abstractions of self and goals, that ultimately lead to goal-directed reaching
behavior. In doing so we have not only shown what those abstractions encode, but have
already capitalized on them for self-supervised motor learning and goal babbling.
We think that these very complementary two cases highlight the generality of our conceptual
and mathematical approach. Of course, it will need more examples in future research to
substantiate this claim.
For that purpose, we think that our work does indeed widely open the door for further investigations
for instance about social learning scenarios \cite{Wrede2012-teleological}
or also other measures of intrinsic motivation \cite{Baranes2013,Merrick2010}.

In particular the case of intrinsic rewards gives rise to interesting 
questions on a more general cognitive science side. In the case of our saliency 
experiment the robot does not start with any particular task semantics. Rather,
its generic information seeking leads to goal-directed behavior. The novelty
in our study is thereby that this behavior is generated and accompanied also
by goal \emph{representations} of what the agent is going for, which is often
seen as the crucial feature of \emph{intentional} action \cite{mcfarland1995opportunity}. 
Are these goals the robot's ``own'' goals? We hope that our
work stimulates discussion around this issue, and can provide more 
detailed answers of how infants become intentional agents 
having their \emph{own} goals, and robots being able to have theirs, too.

\bibliographystyle{IEEEtran}
\bibliography{GoalBabbling,lga}{}

\end{document}